%% file: main.tex
\documentclass{article}

\PassOptionsToPackage{numbers, compress}{natbib}

\usepackage[final]{neurips_2023}

\usepackage[utf8]{inputenc} 
\usepackage[T1]{fontenc}    
\usepackage{booktabs}       
\usepackage{amsfonts}       
\usepackage{nicefrac}       
\usepackage{microtype}      
\usepackage{xcolor}         

\usepackage[pagebackref=true,breaklinks=true,colorlinks,bookmarks=false]{hyperref}      
\usepackage{mysymbols}
\usepackage{soul}
\usepackage{times}
\usepackage{epsfig}
\usepackage{graphicx}
\usepackage{amsmath}
\usepackage{amssymb}
\usepackage{graphicx}
\usepackage[capitalize]{cleveref}
\usepackage{multirow}
\usepackage{wrapfig,lipsum,booktabs}
\usepackage[T1]{fontenc}
\usepackage[utf8]{inputenc}
\usepackage{babel}
\usepackage[font=small,labelfont=bf]{caption}
\usepackage{multirow, makecell}
\usepackage{subcaption}

\usepackage{xspace}
\newcommand*{\eg}{e.g.\@\xspace}
\newcommand*{\ie}{i.e.\@\xspace}
\newcommand*{\etal}{et al.\@\xspace}

\usepackage{tikz}
\usetikzlibrary{arrows.meta}

\def\data{SG-FRONT}
\def\g2d{Graph-to-3D}

\newcommand{\bfx}{{\bf x}}

\newcommand{\bfc}{{\bf c}}

\title{CommonScenes: Generating Commonsense 3D Indoor Scenes with Scene Graph Diffusion}

\makeatletter
\renewcommand*{\@fnsymbol}[1]{}
\makeatother

\author{%
  Guangyao Zhai$^{1,2*}$ \hspace{2mm} Evin P{\i}nar \"{O}rnek$^{1,2*}$ \hspace{2mm} Shun-Cheng Wu$^{1}$ \hspace{2mm} Yan Di$^{1\dag}$\\[0.1em]
  \enspace \, \textbf{Federico Tombari$^{1,3}$} \hspace{2mm} \textbf{Nassir Navab$^{1,2}$} \hspace{2mm} \textbf{Benjamin Busam$^{1,2}$}\AND
  \vspace{-23pt} \\
  \small{\texttt{\{guangyao.zhai,evin.oernek,yan.di\}@tum.de}} \vspace{9pt}\\
  $^1$Technical University of Munich \hspace{2mm} $^2$Munich Center for Machine Learning \hspace{2mm}
  $^3$Google\\[0.8em]
  \href{https://sites.google.com/view/commonscenes}{\texttt{https://sites.google.com/view/commonscenes}}
  \thanks{$^*$The first two authors contributed equally. $^\dag$Corresponding author.}
}

\begin{document}
\maketitle
\input{sections/0_abstract}
\input{sections/1_introduction}

\input{sections/2_related_work}
\input{sections/preliminary}
\input{sections/Methodology}

\input{sections/4_Results}

\input{sections/5_Conclusion}

{\paragraph{\textbf{Acknowledgements}} 
This research is supported by Google unrestricted gift, the China Scholarship Council (CSC), and the Munich Center for Machine Learning (MCML). We are grateful to Google University Relationship GCP Credit Program for supporting this work by providing computational resources. Further, we thank all participants of the perceptual study.}
\newpage
{\small
\bibliographystyle{ieee_fullname}
\bibliography{egbib}
}

\newpage
\appendix
\include{supp}

\end{document}

%% file: sections/0_abstract.tex
\begin{abstract}

Controllable scene synthesis aims to create interactive environments for numerous industrial use cases. 
Scene graphs provide a highly suitable interface to facilitate these applications by abstracting the scene context in a compact manner. 
Existing methods, reliant on retrieval from extensive databases or pre-trained shape embeddings, often overlook scene-object and object-object relationships, leading to inconsistent results due to their limited generation capacity.
To address this issue, we present \emph{CommonScenes}, a fully generative model that converts scene graphs into corresponding controllable 3D scenes, which are semantically realistic and conform to commonsense.
Our pipeline consists of two branches, one predicting the overall scene layout via a variational auto-encoder and the other generating compatible shapes via latent diffusion, capturing global scene-object and local inter-object relationships in the scene graph while preserving shape diversity. 
The generated scenes can be manipulated by editing the input scene graph and sampling the noise in the diffusion model.
Due to the lack of a scene graph dataset offering high-quality object-level meshes with relations, we also construct \emph{SG-FRONT}, enriching the off-the-shelf indoor dataset 3D-FRONT with additional scene graph labels.
Extensive experiments are conducted on SG-FRONT, where \emph{CommonScenes} shows clear advantages over other methods regarding generation consistency, quality, and diversity. %
Codes and the dataset are available on the website.
\end{abstract}

%% file: sections/1_introduction.tex
\section{Introduction}
\label{intro}

\textbf{C}ontrollable \textbf{S}cene \textbf{S}ynthesis (\textbf{CSS}) refers to the process of generating or synthesizing scenes in a way that allows for specific entities of the scene to be controlled or manipulated. Existing methods operate on images~\cite{wang2016gan} or 3D scenes~\cite{Liao2020CVPR} varying by controlling mechanisms from input scene graphs~\cite{sg2im} or text prompts~\cite{brooks2022instructpix2pix}.
Along with the development of deep learning techniques, CSS demonstrates great potential in applications like the film and video game industry~\cite{chen2013study}, augmented and virtual reality~\cite{xiong2021augmented}, and robotics~\cite{Zhao2021,zhai2023sgbot}. For these applications, scene graphs provide a powerful tool to abstract scene content, including scene context and object relationships.\par%

This paper investigates scene graph-based CSS for generating coherent 3D scenes characterized by layouts and object shapes consistent with the input scene graph. To achieve this goal, recent methods propose two lines of solutions. 
The first line of works optimizes scene layouts~\cite{Luo_2020_CVPR} and retrieves objects~\cite{di2023u} from a given database (see Figure~\ref{fig:teaser} (a)). Such retrieval-based approaches are inherently sub-optimal~\cite{uy2021joint} in generation quality due to performance limitations through the size of the database. 
The second solution, \eg, Graph-to-3D~\cite{dhamo2020}, regresses both the layout and shape of the objects for synthesis. However, the shape generation relies on pre-trained shape codes from category-wise auto-decoders, \eg, DeepSDF~\cite{park2019deepsdf}. This semi-generative design (Figure~\ref{fig:teaser} (b)) results in reduced shape diversity in the generated outputs. 
To enhance generation diversity without relying on vast databases, one possible solution is to concurrently predict scene layouts and generate object shapes with text-driven 3D diffusion models\cite{cheng2023sdfusion,li2023diffusionsdf}, where the textual information is obtained from input scene graphs. Yet, in our experiments, we observe that such an intuitive algorithm works poorly since it does not exploit global and local relationship cues among objects encompassed by the graph.\par%
\input{tex/teaser}
In this work, our approach \emph{CommonScenes} exploits global and local scene-object relationships and demonstrates that a fully generative approach can effectively encapsulate and generate plausible 3D scenes without prior databases. %
Given a scene graph, during training, we first enhance it with pre-trained visual-language model features, \eg, CLIP~\cite{radford2021clip}, and bounding box embeddings, incorporating coarse local inter-object relationships into the feature of each node. %
Then, we leverage a triplet-GCN~\cite{Luo_2020_CVPR} based framework to propagate information among objects, learning layouts through global cues and fine local inter-object relationships, which condition the diffusion process~\cite{ldm22} to model the shape distribution as well. %
During inference, each node in the scene graph is enriched with the learned local-to-global context and sequentially fed into the latent diffusion model for each shape generation. %
Figure~\ref{fig:teaser} (c) illustrates that our method effectively leverages relationship encoding to generate commonsense scenes, \ie, arranged plausibly and realistically, exhibiting scene-level consistency while preserving object shape diversity. Furthermore, to facilitate the benchmarking of CSS, we curate a novel indoor scene graph dataset, \emph{\data{}}, upon a synthetic dataset 3D-FRONT~\cite{3dfront}, since no existing indoor scene graph datasets provide high-quality meshes. \data{} comprises around 45K 3D samples with annotated semantic and instance segmentation labels and a corresponding scene graph describing each scene.\par%

Our contributions can be summarized into three points. %
\textbf{First}, we present \emph{CommonScenes}, a fully generative model that converts scene graphs into corresponding 3D scenes using a diffusion model. It can be intuitively manipulated through graph editing. %
\textbf{Second}, \emph{CommonScenes} concurrently models scene layout and shape distribution. It thereby encapsulates both global inter-object relationships and local shape cues. 
 \textbf{Third}, we contribute \emph{SG-FRONT}, 
a synthetic indoor dataset extending 3D-FRONT by scene graphs, thereby contributing graph-conditional scene generation benchmarks.

%% file: tex/teaser.tex
\begin{figure}
    \centering
    \includegraphics[width=\linewidth]{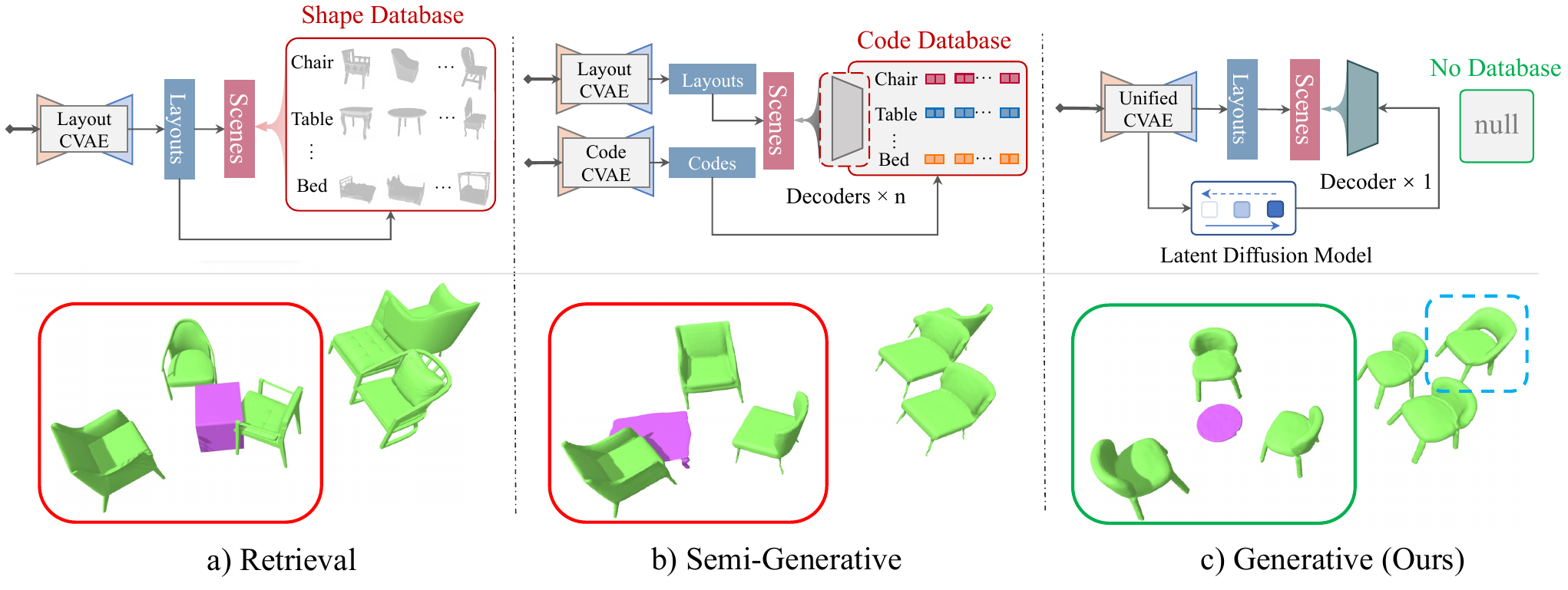}
    \caption{\textbf{Architecture Comparison (Upper Row):} 
    Compared with previous methods, our fully generative model requires neither databases nor multiple category-level decoders. 
    \textbf{Performance Comparison (Bottom Row):} We demonstrate the effectiveness of encapsulating scene-object and object-object relationships.
    The semantic information from the scene graph is \textit{`a table is surrounded by three chairs'}. As highlighted in the rounded rectangles, through the scene-object relationship, our network outperforms other methods by generating a round table and three evenly distributed chairs. Through the object-object relationship, the three chairs are consistent in style. 
    Moreover, our method still preserves the object diversity (blue dashed rectangle).} 
    \label{fig:teaser}
    \vspace{-0.3cm}
\end{figure}

%% file: sections/2_related_work.tex
\section{Related Work}
\paragraph{Scene Graph} %
Scene graphs provide a rich symbolic and semantic representation of the scene using nodes and relationships~\cite{image_retrieval_using_scene_graphs}. They are useful in many 2D-related topics such as image generation~\cite{sg2im,yang2022diffusion}, image manipulation~\cite{dhamo2020}, caption generation~\cite{visual_genome}, visual question answering~\cite{graph_structured_vqa}, and camera localization~\cite{ral2023kabalar}. Quickly after this progress, they are used in following areas: 3D scene understanding~\cite{3DSSG2020,3d_scene_graph_structure_for_unified_semantics,scenegraphfusion,3dssgkim20}, dynamic modeling~\cite{3d_dynamic_scene_graphs}, robotic grounding~\cite{hughes2022hydra,kimera2021,zhai2023sgbot}, spatio-temporal 4D~\cite{4D_OR,kong2020semantic,holistic_4dor}, and controllable scene synthesis~\cite{Li2018grains,zhou2019scenegraphnet,Wang2019PlanITPA,para2021generative,graph2scene2021}.

\paragraph{Indoor 3D Scene Synthesis}
Controllable scene synthesis is extensively explored in the computer graphics community, varying from text-based scene generation~\cite{makeascene} to segmentation map~\cite{park2019SPADE} or spatial layout-based image-to-image translation tasks~\cite{zhaobo2019layout2im}.
Likewise in 3D, several methods generated scenes from images~\cite{factored3dTulsiani17,Nie_2020_CVPR, bahmani2023cc3d,yangguangSST,di2020unifie,ornek2022from2dto3d,Nie_2023_CVPR}, text~\cite{Ma2018language}, probabilistic grammars~\cite{bokeloh2012algebraic,Jiangijcv18,devaranjan2020meta,purkait2020sg}, layouts~\cite{Jyothi_2019_ICCV,tang2023diffuscene}, in an autoregressive manner~\cite{Paschalidou2021NEURIPS,wang2021sceneformer}, or through learning deep priors~\cite{wang2018deep}. %
Another line of work closer to ours is based on graph conditioning~\cite{Li2018grains,zhou2019scenegraphnet,Wang2019PlanITPA,para2021generative}. %
Luo~\etal~\cite{Luo_2020_CVPR} proposed a generative scene synthesis through variational modeling coupled with differentiable rendering. However, their method relies on shape retrieval, which depends on an existing database. %
Graph-to-3D~\cite{graph2scene2021} proposes a joint approach to learn both scene layout and shapes with a scene graph condition. Nevertheless, their object generation relies on a pre-trained shape decoder, limiting the generalizability. %
Unlike previous work, our method generates 3D scenes with the condition over a scene graph, which is trained end-to-end along with content awareness, resulting in higher variety and coherency.

\paragraph{Denoising Diffusion Models} 
A diffusion probabilistic model is a trained Markov chain that generates samples matching data by reversing a process that gradually adds noise until the signal vanishes~\cite{pmlr-v37-sohl-dickstein15}. Diffusion models have quickly gained popularity due to their unbounded, realistic, and flexible generation capacity~\cite{NEURIPS2020_4c5bcfec,song2021scorebased,meng2022sdedit,Karras2022edm,ldm22,di2023ccd}. 
However, studies have identified that the diffusion models lack compositional understanding of the input text~\cite{NEURIPS2022_ec795aea}. 
Several advancements have been introduced to address these limitations. Techniques such as the introduction of generalizable conditioning and instruction mechanisms have emerged~\cite{brooks2022instructpix2pix,controlnet2023}. Moreover, optimizing attention channels during testing has also been explored~\cite{chefer2023attendandexcite,feng2023trainingfree}. 
Recently, a latent diffusion model using a Signed Distance Field (SDF) to represent 3D shapes was proposed contemporaneously at multiple works ~\cite{cheng2023sdfusion,li2023diffusionsdf}, which can be conditioned on a text or a single view image. For the contextual generation conditioned on scene graphs, methods have converted triplets into the text to condition the model~\cite{feng2023trainingfree,yang2022diffusion}, where Yang~\etal~\cite{yang2022diffusion} proposed a graph conditional image generation based on masked contrastive graph training. To the best of our knowledge, we are the first to leverage both areas of scene graph and latent diffusion for end-to-end 3D scene generation.

%% file: sections/preliminary.tex
\section{Preliminaries}

\paragraph{Scene Graph}
A scene graph, represented as $\mathcal{G} = (\mathcal{V}, \mathcal{E})$, is a structured representation of a visual scene where $\mathcal{V} = \{ v_{i}~|~i \in \{1,  \ldots, N \} \}$ denotes the set of vertices (object nodes) and $\mathcal{E} = \{e_{i \to j}~|~i,j \in \{1, \ldots, N\}, i \neq j \}$ represents the set of directed edge from node $v_{i}$ to $v_{j}$ . 
Each vertex $v_{i}$  is categorized through an object class $c^{node}_{i} \in \mathcal{C}^{node}$, where $\mathcal{C}^{node}$ denotes the set of object classes.
The directed edges in $\mathcal{E}$ capture the relationships between objects in terms of both semantic and geometric information. Each edge $e_{i\to j}$ has a predicate class $c^{edge}_{i\to j} \in \mathcal{C}^{edge}$, where $\mathcal{C}^{edge}$ denotes the set of edge predicates. 
These relationships can incorporate various aspects, such as spatial locations (\eg, \texttt{left/right, close by}) or object properties (\texttt{bigger/smaller}).
To facilitate subsequent processing and analysis, each node $v_i$ and edge $e_{i\to j}$ are typically transformed into learnable vectors $o_i$ and $\tau_{i\to j}$, respectively, through embedding layers, as shown in Figure~\ref{fig:contextual}.A.

\paragraph{Conditional Latent Diffusion Model} Diffusion models learn a target distribution by reversing a progressive noise diffusion process modeled by a fixed Markov Chain of length $T$~\cite{NEURIPS2020_4c5bcfec,song2021scorebased}. Fundamentally, given a sample $\bfx_{t}$ from the latent space, gradual Gaussian Noise is added with a predefined scheduler $\bfx_{t}, t \in \{ 1, \ldots, T\}$~\cite{NEURIPS2020_4c5bcfec}. Then, the denoiser $\varepsilon_\theta$, typically a UNet~\cite{unet}, is trained to recover denoising from those samples. The recently introduced Latent Diffusion Models (LDMs)~\cite{ldm22} reduce the computational requirements by learning this distribution in a latent space established by a pre-trained VQ-VAE~\cite{oord2017vqvae} instead of directly processing the full-size input. The popular usage of LDM is conditional LDM, which allows the generation to obey the input cue $\bfc_i$~\cite{ldm22}. The training objective can be simplified to
\begin{equation}
\label{eqn:sdf}
    \mathcal{L}_{LDM} = 
    \mathbb{E}_{\bfx,\varepsilon\sim\mathcal{N}(0,1),t} 
    \left [ || \varepsilon - \varepsilon_\theta(\bfx_t, t, \bfc_i) ||_2^2 \right ],
\end{equation}
where $\bfc_i$ denotes a conditioning vector corresponding to the input sample $i$ fed into $\varepsilon_\theta$. 

%% file: sections/Methodology.tex
\section{Method}
\input{tex/contextual}
\paragraph{Overview}
Given a semantic scene graph, our approach endeavors to generate corresponding 3D scenes conforming to commonsense.
We employ a dual-branch network starting with a contextual encoder $E_c$, as shown in Figure~\ref{fig:pipeline}.
The two branches, referred to as the \textit{Layout Branch} and the \textit{Shape Branch}, function simultaneously for layout regression and shape generation. In Sec.~\ref{sge}, we first illustrate how a scene graph evolves to a \textbf{B}ox-enhanced \textbf{C}ontextual \textbf{G}raph (BCG) with features from pre-trained visual-language model CLIP~\cite{radford2021clip} and bounding box parameters (Figure~\ref{fig:contextual}). Then, we show how BCG is encoded by $E_c$ and manipulated by the graph manipulator (Figure~\ref{fig:pipeline}.A, B and C). In Sec.~\ref{lb}, we introduce the layout branch for layout decoding, and in Sec.~\ref{sb}, we introduce the shape branch for shape generation. Finally, we explain the joint optimization in Sec.~\ref{lst}.

\subsection{Scene Graph Evolution}
\label{sge}
\paragraph{Contextual Graph} As shown in Figure~\ref{fig:contextual}.A and B, we incorporate readily available prompt features from CLIP~\cite{radford2021clip} as semantic anchors, capturing coarse inter-object information, into each node and edge of the input graph to conceptualize it as \emph{Contextual Graph} $\mathcal{G}_c$ with 
$p_{i} = E_{\text{CLIP}}(c^{node}_{i})$ for objects and $ p_{i\to j} = E_{\text{CLIP}}(c^{node}_{i} \boxplus c^{edge}_{i\to j} \boxplus c^{node}_{j})$ for edges. Here $E_{\text{CLIP}}$ is the pre-trained and frozen text encoder in~\cite{radford2021clip}, $\boxplus$ denotes the aggregation operation on prompts including subject class $c^{node}_{i}$, predicate $c^{edge}_{i\to j}$, and object class $c^{node}_{j}$. Thereby, the embeddings of $\mathcal{G}_{c}=(\mathcal{V}_{c},\mathcal{E}_{c})$ is formalized as,
\begin{small} 
\begin{equation}
  \mathcal{F}_{\mathcal{V}_{c}} = \{ f_{v_c}^{i}=(p_{i}, o_{i})~|~i \in \{1, \ldots, N \} \}, ~
  \mathcal{F}_{\mathcal{E}_{c}} = \{f_{e_c}^{i\to j}=(p_{i\to j}, \tau_{i\to j})~|~i,j \in \{1, \ldots, N \}\},
\end{equation}
\end{small}%
where $\mathcal{F}_{\mathcal{V}_{c}}$ represents the set of object features and $\mathcal{F}_{\mathcal{E}_{c}}$ the edge features.

\paragraph{Box-Enhanced Contextual Graph} 
In training, we enrich each node in the contextual graph by using ground truth bounding boxes parameterized by box sizes $s$, locations $t$, and the angular rotations along the vertical axis $\alpha$, yielding a BCG, as shown in Figure~\ref{fig:contextual}.C. 
Thereby, the node embeddings of BCG are represented by $\mathcal{F}_{\mathcal{V}_c}^{(b)}=\{ f_{v_c}^{(b)i}=(p_{i}, o_{i}, b_{i})~|~i \in \{1, \ldots, N \}\}$, where $b_{i}$ is obtained by encoding $(s_i, t_i, \alpha_i)$ with MLPs. 
Note that BCG is only used during training, i.e., bounding box information is not needed during inference.

\paragraph{Graph Encoding} 
BCG is encoded by the subsequent triplet-GCN-based contextual encoder $E_{c}$, which together with the layout decoder $D_l$ in Sec.~\ref{lb} construct a Conditional Variational Autoencoder (CVAE)~\cite{Luo_2020_CVPR}. The encoding procedure is shown in Figure~\ref{fig:pipeline}.A, B and C. Given a BCG, during training, we input the embeddings $\mathcal{F}_{\mathcal{V}_c}^{(b)}$ and $\mathcal{F}_{\mathcal{E}_{c}}$ into $E_{c}$ to obtain an \textit{Updated Contextual Graph} with node and edge features represented as $(\mathcal{F}_{\mathcal{V}_c}^{(z)}, \mathcal{F}_{\mathcal{E}_{c}})$, which is also the beginning point of the inference route. 
Each layer of $E_{c}$ consists of two sequential MLPs $\{g_1, g_2\}$, where $g_1$ performs message passing between connected nodes and updates the edge features,  $g_2$ aggregates features from all connected neighbors of each node and update its features, as shown in the follows:
\begin{equation}
    \begin{aligned}
    (\psi_{v_{i}}^{l_g}, \phi_{e_{i\to j}}^{l_g+1}, \psi_{v_{j}}^{l_g}) &= g_1(\phi_{v_{i}}^{l_g}, \phi_{e_{i\to j}}^{l_g}, \phi_{v_{j}}^{l_g}), \quad l_g = 0, \ldots, L - 1, \\
    \phi_{v_{i}}^{l_g+1} &= \psi_{v_{i}}^{l_g} + g_{2}\Big( \text{AVG}\big( \psi_{{v}_{j}}^{l_g}~|~v_j \in N_{\mathcal{G}}(v_{i}) \big) \Big),
    \label{eq:gcn}
    \end{aligned}
\end{equation}
where $l_g$ denotes a single layer in $E_c$ and $N_{\mathcal{G}}(v_{i})$ includes all the connected neighbors of $v_{i}$. 
AVG refers to average pooling.
We initialize the input embeddings of layer $0$ as the features from the \textit{Updated Contextual Graph}, $(\phi_{v_{i}}^{0}, \phi_{e_{i\to j}}^{0}, \phi_{v_{j}}^{0})=(f_{v_c}^{(b)i}, f_{e_c}^{i\to j}, f_{v_c}^{(b)j})$. 
The final embedding $\phi_{v_{i}}^{L}$ is used to model a joint layout-shape distribution $Z$, parameterized with the $N_c$-dimensional Gaussian distribution $Z\sim N(\mu, \sigma)$, where $\mu,\sigma \in \mathbb{R}^{N_c}$,  predicted by two separated MLP heads.
Thereby, $Z$ is modeled through minimizing:
\begin{equation}
 \mathcal{L}_{KL} = D_{KL}\left(E^{\theta}_{c}(z | x, \mathcal{F}_{\mathcal{V}_c}^{(b)}, \mathcal{F}_{\mathcal{E}_{c}})~||~p(z | x)\right),
\label{kl}
\end{equation}
where $D_{KL}$ is the Kullback-Liebler divergence measuring the discrepancy between $Z$ and the posterior distribution $p(z | x)$ chosen to be standard $N(z~|~0, 1)$. 
We sample a random vector $z_i$  from $Z$ for each node, update node feature $\mathcal{F}^{(z)}_{\mathcal{V}_c}$ by concatenating $\{z_i, p_i, o_i\}$, keep 
edge features unchanged. 

\subsection{Layout Branch}
\label{lb}
\input{tex/pipeline}
 As shown in Figure~\ref{fig:pipeline}.D, a decoder $D_{l}$, which is another triplet-GCN, generates 3D layout predictions upon updated embeddings. In this branch, $E_c$ and $D_l$ are jointly optimized by Eq.~\eqref{kl} and a bounding box reconstruction loss: 
\begin{equation}
    \mathcal{L}_{layout} =  \frac{1}{N} \sum_{i=1}^{N}(|s_i - \hat{s}_i|_1+ |t_i - \hat{t}_i|_1 - \sum_{\lambda=1}^{\Lambda} \alpha^{\lambda}_i \log \hat{\alpha}^{\lambda}_i),
\end{equation}
where $\hat{s_i}, \hat{t_i}, \hat{\alpha_i}$ denote the predictions of bounding box size $s$, locations $t$ and angular rotations $\alpha$, respectively.
$\lambda$ is the rotation classification label.
We partition the rotation space into $\Lambda$ bins, transforming the rotation regression problem into a classification problem.

\subsection{Shape Branch}
\label{sb}
As shown in Figure~\ref{fig:pipeline}.E, in parallel to the \textit{Layout Branch}, we introduce \textit{Shape Branch} to generate shapes for each node in the given graph, and represent them by SDF.

\paragraph{Relation Encoding} 
The core idea of our method is to exploit global scene-object and local object-object relationships to guide the generation process.
Hence, besides incorporating CLIP features for coarse local object-object relationships, as introduced in \cref{sge}, we further design a relation encoder $E_r$, based on triplet-GCN as well, to encapsulate global semantic cues of the graph into each node and propagate local shape cues between connected nodes.
Specifically, $E_r$ operates on the learned embeddings $(\mathcal{F}_{\mathcal{V}_c}^{(z)}, \mathcal{F}_{\mathcal{E}_{c}})$ of the \textit{Updated Contextual Graph}, and updates the feature of each node with local-to-global semantic cues, yielding node-relation embeddings for subsequent diffusion-based shape generation.

\paragraph{Shape Decoding} 
We use an LDM conditioned on node-relation embedding to model the shape-generation process. 
In terms of shape representation, we opt for truncated SDF (TSDF)~\cite{curless1996volumetric} around the surface of the target object within a voxelized space $S \in \mathbb{R}^{D \times D \times D}$. 
Following the LDM, the dimension $D$ of TSDF can be reduced by training a VQ-VAE~\cite{oord2017vqvae} as a shape compressor to encode the 3D shape into latent dimensions $\bfx \in \mathbb{R}^{d \times d \times d}$, where $d << D$ is built upon a discretized codebook. 
Then a forward diffusion process adds random noise to the input shape $\bfx_0$ transferring to $\bfx_T$, upon which we deploy a 3D-UNet~\cite{3dunet2016} $\varepsilon_\theta$ (Figure~\ref{fig:pipeline}.E) to denoise the latent code back to $\bfx_0$ according to DDPM model by Ho \etal~\cite{NEURIPS2020_4c5bcfec}. The denoiser is conditioned on node-relation embeddings to intermediate features of 3D UNet via the cross-attention mechanism. 
Finally, the decoder of VQ-VAE generates the shape $S'$ from the reconstructed $\bfx_0$. 
For the denoising process at the timestep $t$, the training objective is to minimize: 
\begin{equation}
    \mathcal{L}_{shape} = 
    \mathbb{E}_{{\bfx},\varepsilon\sim\mathcal{N}(0,1),t} 
    \left [ || \varepsilon - \varepsilon_\theta(\bfx_t, t,  E_{r}(\mathcal{F}_{\mathcal{V}_c}^{(z)}, \mathcal{F}_{\mathcal{E}_{c}})||_2^2 \right ],
\end{equation}
where the evidence lower bound between the sampled noise $\varepsilon$ and the prediction conditioned on the contextual relation embedding extracted from $E_r$ is optimized. At test time, a latent vector is randomly sampled from $\mathcal{N}(0,1)$ and progressively denoised by 3D-Unet to generate the final shape. Each shape within the layout is populated based on per-node conditioning, ultimately producing a plausible scene. Compared to prior work, our design can bring more diverse shape generation by taking advantage of the diffusion model architecture, as shown in experiments. 

\subsection{Layout-Shape Training} 
\label{lst}
Our pipeline is trained jointly in an end-to-end fashion, allowing the \textit{Layout Branch} and \textit{Shape Branch} to optimize each other by sharing latent embeddings $(\mathcal{F}_{\mathcal{V}_c}^{(z)}, \mathcal{F}_{\mathcal{E}_{c}})$, coming out of $E_c$. 
The final optimization loss is the combination of scene distribution modeling, layout generation, and shape generation:
\begin{equation}
    \mathcal{L} = \lambda_1 \mathcal{L}_{KL} + \lambda_2 \mathcal{L}_{layout} + \lambda_3 \mathcal{L}_{shape},
\end{equation}
where $\lambda_1$, $\lambda_2$ and $\lambda_3$ are weighting factors. Further insights on accomplishing the batched training are provided in Supplementary Material.

%% file: tex/contextual.tex
\begin{figure}
    \centering
    \includegraphics[width=1.0\linewidth]{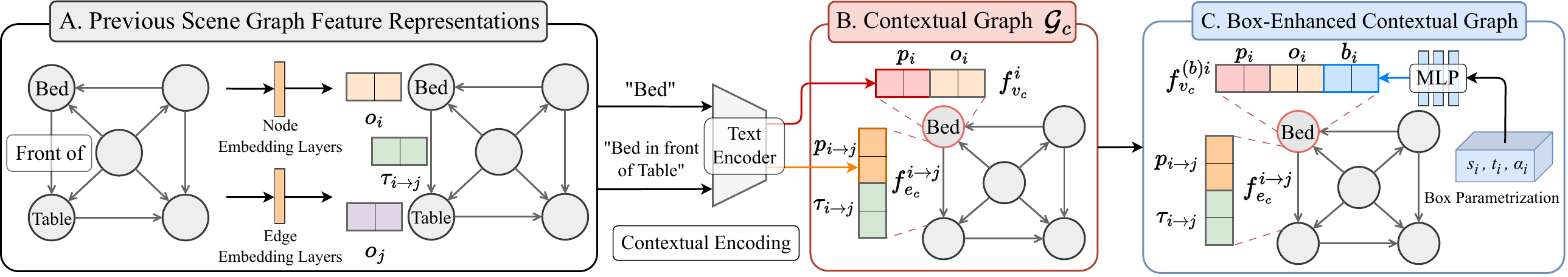}
    \caption{\textbf{Scene Graph Evolution.} 
    Take the features of two nodes \textit{Bed $(o_i)$, Table $(o_j)$} and the linked edge \textit{In front of $(\tau_{i\to j})$} as an example, where $(o_i, o_j), \tau_{i\to j}$ are embedded learnable node features and the edge feature, respectively.
    We enhance the node and edge features with CLIP feature $p_i, p_{i\to j}$ to obtain \textit{B. Contextual Graph}. 
    Then, we parameterize the ground truth bounding box $b_i$ to the node to further build \textit{C. Box-Enhanced Contextual Graph} with node and edge feature represented as $f_{v_c}^{(b)i}=\{p_i, o_i, b_i\}, f_{e_c}^{i\to j}=\{p_{i\to j}, \tau_{i\to j}\}$. 
    }
    \label{fig:contextual}
    \vspace{-0.3cm}
\end{figure}

%% file: tex/pipeline.tex
\begin{figure}
    \centering
    \includegraphics[width=1.0\linewidth]{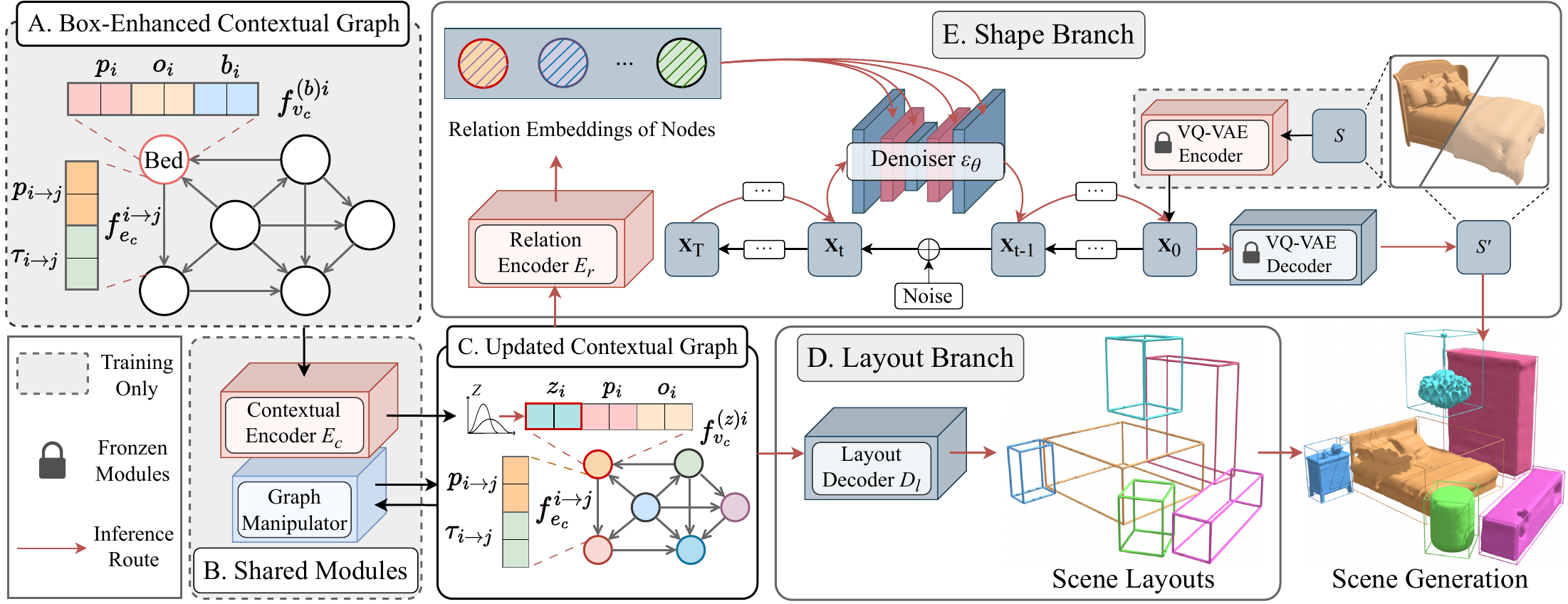}
    \caption{\textbf{Overview of CommonScenes.} Our pipeline consists of shared modules and two collaborative branches \textit{Layout Branch} and \textit{Shape Branch}. 
    Given a BCG (Figure~\ref{fig:contextual}.C), we first feed it into $E_c$, yielding a joint layout-shape distribution $Z$. 
    We sample $z_i$ from $Z$ for each node, obtaining concatenated feature $\{z_i, p_i, o_i\}$ with CLIP feature $p_i$ and self-updated feature $o_i$.
    A graph manipulator is then optionally adopted to manipulate the graph for data augmentation.
    Next, the updated contextual graph is fed into the layout branch and shape branch for layout regression and shape generation respectively.
    In the shape branch, we leverage $E_r$ to encapsulate global scene-object and local object-object relationships into graph nodes, which are then conditioned to $\varepsilon_\theta$ in LDM via cross-attention mechanism to generate $\bfx_0$ back in $T$ steps.  
    Finally, a frozen shape decoder (VQ-VAE) reconstructs $S'$ using $\bfx_0$.  
    The final scene is generated by fitting $S'$ to layouts.
    }
    \label{fig:pipeline}
    \vspace{-0.3cm}
\end{figure}

%% file: sections/4_Results.tex
\section{Experiments}

\input{tex/qual_comp}
\paragraph{SG-FRONT Dataset} Due to the lack of scene graph datasets also providing high-quality object meshes, we construct \emph{SG-FRONT}, a set of well-annotated scene graph labels, based on a 3D synthetic dataset 3D-FRONT~\cite{3dfront} that offers professionally decorated household scenarios.
The annotation labels can be grouped into three categories \textbf{\textit{spatial/proximity, support, and style}}. 
The spatial relationships are based on a series of relationship checks between the objects, which control the object bounding box location, e.g., \texttt{left/right}, the comparative volumes, e.g., \texttt{bigger/smaller}, and heights, e.g., \texttt{taller/shorter}. 
The support relationships include directed structural support, e.g., \texttt{close by}, \texttt{above}, and \texttt{standing on}.
Thresholds for these labels are iterated and set by the annotators.
Finally, style relationships include the object attributes to assign labels, namely \texttt{same material as}, \texttt{same shape as}, and \texttt{same super-category as}.
SG-FRONT contains around 45K samples from three different indoor scene types, covering 15 relationship types densely annotating scenes.
More details are provided in the Supplementary Material.

\paragraph{Implementation Details}
We conduct the training, evaluation, and visualization of \emph{CommonScenes} on a single NVIDIA A100 GPU with 40GB memory. We adopt the AdamW optimizer with an initial learning rate of 1e-4 to train the network in an end-to-end manner.
We set $\{\lambda_1, \lambda_2, \lambda_3\} = \{1.0, 1.0, 1.0\}$ in all our experiments. 
$N_c$ in distribution $Z$ is set to 128 and TSDF size $D$ is set as 64.
We provide more details in the Supplementary Material.

\paragraph{Evaluation Metrics}
To measure the \textbf{fidelity and diversity} of generated scenes, we employ the commonly adopted Fréchet Inception Distance (FID)~\cite{fid2017} \& Kernel Inception Distance (KID)~\cite{kid2018} metrics \cite{Paschalidou2021NEURIPS}. 
We project scenes onto a bird's-eye view, excluding lamps to prevent occlusion and using object colors as semantic indicators. To measure the \textbf{scene graph consistency}, we follow the scene graph constraints~\cite{graph2scene2021}, which measure the accuracy of a set of relations on a generated layout. We calculate the spatial \texttt{left/right, front/behind, smaller/larger, taller/shorter} metrics, as well as more difficult proximity constraints \texttt{close by} and \texttt{symmetrical}. 
To measure the \textbf{shape consistency}, we test dining rooms, a typical scenario in which tables and chairs should be in suits according to the commonsense decoration. We identify matching dining chairs and tables in 3D-FRONT using the same CAD models and note their instance IDs. This helps us determine which entities belong to the same sets. We calculate Chamfer Distance (CD) of each of the two objects in each set after generation. To measure the \textbf{shape diversity}, we follow~\cite{graph2scene2021} to generate each scene 10 times and evaluate the change of corresponding shapes using CD. We illustrate the concepts of diversity and consistency in Figure~\ref{fig:objs}. We also report MMD, COV, and 1-NNA following~\cite{yang2019pointflow} for \textbf{the object-level evaluation} in the Supplementary Materials.

\begin{table*}[t!]
\centering
    \scalebox{0.85}{
    \begin{tabular}{l  c  c c | c c | c c | c c }
     \toprule 
        \multirow{2}{*}{Method} & Shape & \multicolumn{2}{c}{Bedroom} & \multicolumn{2}{c}{Living room} & \multicolumn{2}{c}{Dining room} & \multicolumn{2}{c}{All}
        \\ & Representation & FID& KID & FID& KID& FID& KID & FID& KID \\
    \midrule 
        3D-SLN~\cite{Luo_2020_CVPR} & \multirow{4}{*}{Retrieval} & 57.90 & 3.85 & 77.82 & 3.65 & 69.13 & 6.23 & 44.77 & 3.32\\
        Progressive~\cite{graph2scene2021} &  & 58.01 & 7.36 & 79.84 & 4.24 & 71.35 & 6.21 & 46.36 & 4.57 \\
        Graph-to-Box~\cite{graph2scene2021} &  & 54.61 & 2.93 &  78.53 & 3.32 & 67.80 & 6.30 & 43.51 & 3.07 \\
        \textbf{Ours} w/o SB &  & \textbf{52.69} & \textbf{2.82} & \textbf{76.52} & \textbf{2.08} & \textbf{65.10} & \textbf{6.11} & \textbf{42.07} & \textbf{2.23}\\
    \midrule 
    Graph-to-3D~\cite{graph2scene2021} & DeepSDF~\cite{park2019deepsdf} & 63.72 & 17.02 & 82.96 & 11.07 & 72.51 & 12.74 & 50.29 & 7.96\\
    Layout+txt2shape  & SDFusion~\cite{cheng2023sdfusion} &  68.08 & 18.64& 85.38 & 10.04 & \textbf{64.02} & \textbf{5.08} & 50.58 & 8.33\\
    \textbf{Ours} & rel2shape & \textbf{57.68} & \textbf{6.59} & \textbf{80.99} & \textbf{6.39} & 65.71 & 5.47 & \textbf{45.70} & \textbf{3.84}\\
    \bottomrule
    \end{tabular}
    }
\caption{\textbf{Scene generation realism} as measured by FID and KID $(\times 0.001)$ scores at $256^{2}$ pixels between the top-down rendering of generated and real scenes (lower is better). Two main rows are separated with respect to the reliance on an external shape database for retrieval. “Ours w/o SB” refers to ours without the shape branch.}
\label{tab:fidkid}
\end{table*}

\paragraph{Compared Baselines}
We include three types of baseline methods. \textbf{First}, three retrieval-based methods, \ie, a layout generation network \emph{3D-SLN}~\cite{Luo_2020_CVPR}, a progressive method to add objects one-by-one designed in~\cite{graph2scene2021}, and \emph{Graph-to-Box} from~\cite{graph2scene2021}. \textbf{Second}, a semi-generative SOTA method \emph{Graph-to-3D}. \textbf{Third}, an intuitive method called \emph{layout+txt2shape} that we design according to the instruction in Sec.~\ref{intro}, stacking a layout network and a text-to-shape generation model in series, and  it is a fully generative approach with text only. All baseline methods are trained on the SG-FRONT dataset following the public implementation and training details. 

\subsection{Graph Conditioned Scene Generation}
\begin{figure}[t!]
\centering
\begin{minipage}[b]{0.48\textwidth}
\centering
\includegraphics[width=0.75\linewidth]{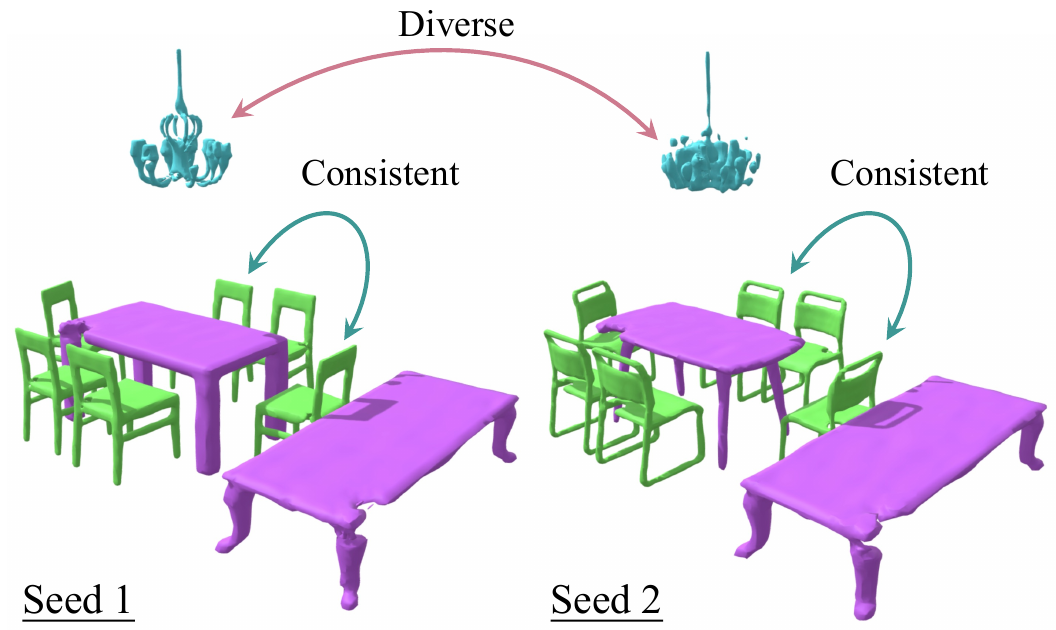}
\captionof{figure}{\textbf{Consistency co-exists with diversity in different rounds.} Our generated objects show diversity when activated twice while preserving the shape consistency within the scene (chairs in a suit).\label{fig:objs}}
\end{minipage}
\hfill
\begin{minipage}[b]{0.48\textwidth}
\centering
\scalebox{0.8}{
\begin{tabular}{l cc|cc}
    \toprule
         \multirowcell{2}{Method} & \multicolumn{2}{c|}{Consistency $\downarrow$} & \multicolumn{2}{c}{Diversity $\uparrow$} \\
         \cmidrule{2-5}
         & Chair & Table & Chair & Table\\
     \midrule
         Graph-to-Box~\cite{graph2scene2021} & 10.42  & 50.66 & 0.53 & 0.89\\
         Graph-to-3D~\cite{graph2scene2021} & 2.49  & 11.74 & 1.43 & 0.93\\
         Ours & \textbf{1.96}  & \textbf{9.04} & \textbf{30.04} & \textbf{10.53}\\
     \bottomrule
    \end{tabular}
    }
\captionof{table}{\textbf{Consistency and diversity in the dining rooms.} The object shapes related with \textit{same as} within a scene are consistent as indicated by low CD values ($\times 0.001$), whereas the shapes across different runs have high diversity,  per high KL divergence.\label{tab:objs}}
\end{minipage}
\vspace{-0.4cm}
\end{figure}

\paragraph{Qualitative results} 
The generated scenes from different baselines are shown in Figure~\ref{fig:qual_comp}:
(a) Graph-to-Box retrieval, (b) Graph-to-3D, (c) layout+txt2shape, and (d) ours. 
It can be seen that our method has better shape consistency and diversity. 
Specifically, in terms of object shape quality, retrieval baseline scenes look optimal since they are based on real CAD models. Yet, the layout configuration is poor, and the retrieved object styles are not plausible, e.g., dining chairs and nightstands are not in a suit, respectively, in the same rooms. Graph-to-3D improves this by learning coherent scenes. However, the object styles are not conserved within scene styles, providing unrealistic scene-object inconsistency, e.g., in the bedroom, the height of the chair does not match the style and the height of the table. Subsequently, (c) improves the layout, which can be seen in the living room area, but again falls back on the generated shape quality and shares the same problem with Graph-to-Box on the poor object-object consistency. In contrast, CommonScenes can capture diverse environments considering both generation consistency and stylistic differences with respect to each scene. In the living room, the scene graph input requires that Table No.3 and Table No.4 should stand close to each other. Graph-to-Box and Graph-to-3D fail to achieve the goal, which is also reflected in Table~\ref{tab:generation_3dfront} and Table~\ref{tab:mani} that these kinds of methods cannot handle \texttt{close by} relation. We show more results in Supplementary Material.

\paragraph{Quantitative results} We provide the FID/KID scores in \cref{tab:fidkid} separated as average over three room types. Our method establishes the best results among the free-shape representation methods by improving the Graph-to-3D by approximately $10\%$ on FID and $40\%$ on KID on average. On the other hand, most of
 the \textit{layout+txt2shape} results are worse than the previous state-of-the-art methods, proving the requirement of learning layout and shape jointly.
Notably, the retrieval-based methods overall have better scores than those of free shapes since the retrieved objects align well with the test database. Further on, among the retrieval baselines, ours without the diffusion shape branch (\textit{Ours w/o SB}) surpasses the prior works, indicating the benefit of contextual graph representation. Additionally, on the generation consistency, we can observe in \cref{tab:objs} that our method can reflect the edge relationship in the generated shapes directly, indicated by significant improvement in the consistency score. By leveraging the latent diffusion model, our diversity is significantly improved compared with other methods. We provide more results on diversity and also show perceptual study on the evaluation of consistency and diversity in the Supplementary Material.

In terms of scene graph constraints, we present the results as easy and hard metrics in \cref{tab:generation_3dfront}. On easy metrics, our methods (Ours, Ours w/o SB) are either better than or on par with other methods. While on hard metrics, they are superior to others. It shows that the spatial relationships could be learned easier than the proximity (\texttt{close by}) and commonsense (\texttt{symmetrical}) based ones. The improvement over the \texttt{symmetrical} constraint is particularly evident, as our method surpasses the state-of-the-art (SOTA) by $10\%$. This demonstrates that the integrated layout and shape cues are crucial for learning these contextual configurations.

\begin{table*}[t!]
    \centering
    \scalebox{0.72}{
    \begin{tabular}{l c cccc | cc }
    \toprule 
    \multirow{2}{*}{Method} & \multirowcell{2}{Shape\\Representation} & \multicolumn{4}{c}{Easy} & \multicolumn{2}{|c}{\textbf{Hard$*$}} \\
    \cmidrule{3-6}\cmidrule{7-8}
     &   & left / right& front / behind & smaller / larger& taller / shorter& \textbf{close by$*$} & \textbf{symmetrical$*$} \\
    \midrule 
        3D-SLN~\cite{Luo_2020_CVPR} & \multirow{4}{*}{Retrieval} & 0.97 & 0.99 & 0.95 & 0.91 & 0.72 & 0.47 \\
        Progressive~\cite{graph2scene2021} &  & 0.97 & 0.99 & 0.95 & 0.82 & 0.69 & 0.46 \\ 
        Graph-to-Box~\cite{graph2scene2021} &  & \textbf{0.98} & 0.99 & 0.96 & \textbf{0.95} & 0.72 & 0.45 \\
        \textbf{Ours} w/o SB &  & \textbf{0.98} & 0.99 & \textbf{0.97} & \textbf{0.95} & 0.74 & \textbf{0.63} \\
    \midrule 
        Graph-to-3D~\cite{graph2scene2021} & DeepSDF~\cite{park2019deepsdf} & \textbf{0.98} & 0.99 & \textbf{0.97} & \textbf{0.95} & 0.74 & 0.57\\
        \textbf{Ours} & rel2shape & \textbf{0.98} & \textbf{1.00} & \textbf{0.97} & \textbf{0.95} & \textbf{0.77} & 0.60\\
    \bottomrule
    \end{tabular}
    }
    \caption{Scene graph constrains on the \textbf{generation} task (higher is better). The total accuracy is computed as the mean over the individual edge class accuracy to minimize class imbalance bias.}
    \label{tab:generation_3dfront}
    \vspace{-0cm}
\end{table*}

\begin{table*}[t!]
    \centering
    \scalebox{0.68}{
    \begin{tabular}{l c c cccc | cc}
    \toprule
    \multirow{2}{*}{Method} & \multirow{2}{*}{\makecell{Shape \\ Representation}} &\multirow{2}{*}{Mode}& \multicolumn{4}{c}{Easy} & \multicolumn{2}{|c}{\textbf{Hard$*$}}\\
    \cmidrule{4-7}\cmidrule{8-9}
     &  &  & left /right & front / behind& smaller / larger& taller / shorter& \multirow{1}{*}{\textbf{close by$*$}} & \multirow{1}{*}{\textbf{symmetrical$*$}}   \\
        
    \midrule 
        3D-SLN~\cite{Luo_2020_CVPR} & \multirow{4}{*}{Retrieval} & \multirow{7}{*}{change} & 0.89 & 0.90 & 0.55 & 0.58 & 0.10 & 0.09  \\
        Progressive~\cite{graph2scene2021} & && 0.89 & 0.89 & 0.52 & 0.55 & 0.08 & 0.09  \\
        Graph-to-Box~\cite{graph2scene2021} & & & \textbf{0.91} & 0.91 & \textbf{0.86} & 0.91 & 0.66 & 0.53  \\
        \textbf{Ours} w/o SB & & & \textbf{0.91} & \textbf{0.92} & \textbf{0.86} & \textbf{0.92} & \textbf{0.70} & 0.53  \\
        \cmidrule{1-2} \cmidrule{4-9}
        Graph-to-3D~\cite{graph2scene2021} &  DeepSDF~\cite{park2019deepsdf} & & \textbf{0.91} & \textbf{0.92} & \textbf{0.86} & 0.89 & 0.69 & 0.46 \\
        \textbf{Ours} & rel2shape & & \textbf{0.91} & \textbf{0.92} & \textbf{0.86} & 0.91 & 0.69 & \textbf{0.59} \\
        \midrule \midrule
        3D-SLN~ \cite{Luo_2020_CVPR} & \multirow{4}{*}{Retrieval} & \multirow{7}{*}{addition} & 0.92 & 0.92 & 0.56 & 0.58 & 0.05 & 0.05 \\
        Progressive & &  & 0.92 & 0.91 & 0.53 &  0.54 & 0.02 & 0.06 \\
        Graph-to-Box~\cite{graph2scene2021} &  & & 0.94 & 0.93 & 0.90 & 0.94 & 0.67 & 0.58 \\
        \textbf{Ours} w/o SB & & & \textbf{0.95} & \textbf{0.95} & 0.90 & 0.94 & \textbf{0.73} & \textbf{0.63}\\
        \cmidrule{1-2} \cmidrule{4-9}
        Graph-to-3D~\cite{graph2scene2021} &  DeepSDF~\cite{park2019deepsdf} & &  0.94 & \textbf{0.95} & \textbf{0.91} & 0.93 & 0.63 & 0.47\\
        \textbf{Ours} & rel2shape & & \textbf{0.95} & \textbf{0.95} & \textbf{0.91} & \textbf{0.95} & 0.70 & 0.61\\
    \midrule
    \end{tabular}}
    \caption{Scene graph constraints on the \textbf{manipulation} task (higher is better). The total accuracy is computed as the mean over the individual edge class accuracy to minimize class imbalance bias. Top: Relationship change mode. Bottom: Node addition mode.}
    \label{tab:mani}
    \vspace{-0.3cm}
\end{table*}

\subsection{Scene Manipulation}

We demonstrate the downstream application of scene manipulation on our method and compare it with the aforementioned methods on scene graph constraints in Table \ref{tab:mani}. Our methods have the highest total score in both relation change and object addition modes. In both modes, the results are on par with Graph-to-3D on easy relations, whereas there is a major improvement in the hard ones. Further, within the retrieval methods, Ours w/o SB improves over the baselines in the addition mode and is on par in the change mode. It could be argued that changing could be more difficult since it could alter multiple objects, whereas object insertion has less effect on the overall scene. We show some qualitative manipulation examples in the Supplementary Material.

\subsection{Ablations}
\begin{figure}[h!]
\centering
\begin{minipage}[b]{0.5\textwidth}
\centering
\includegraphics[width=0.9\linewidth]{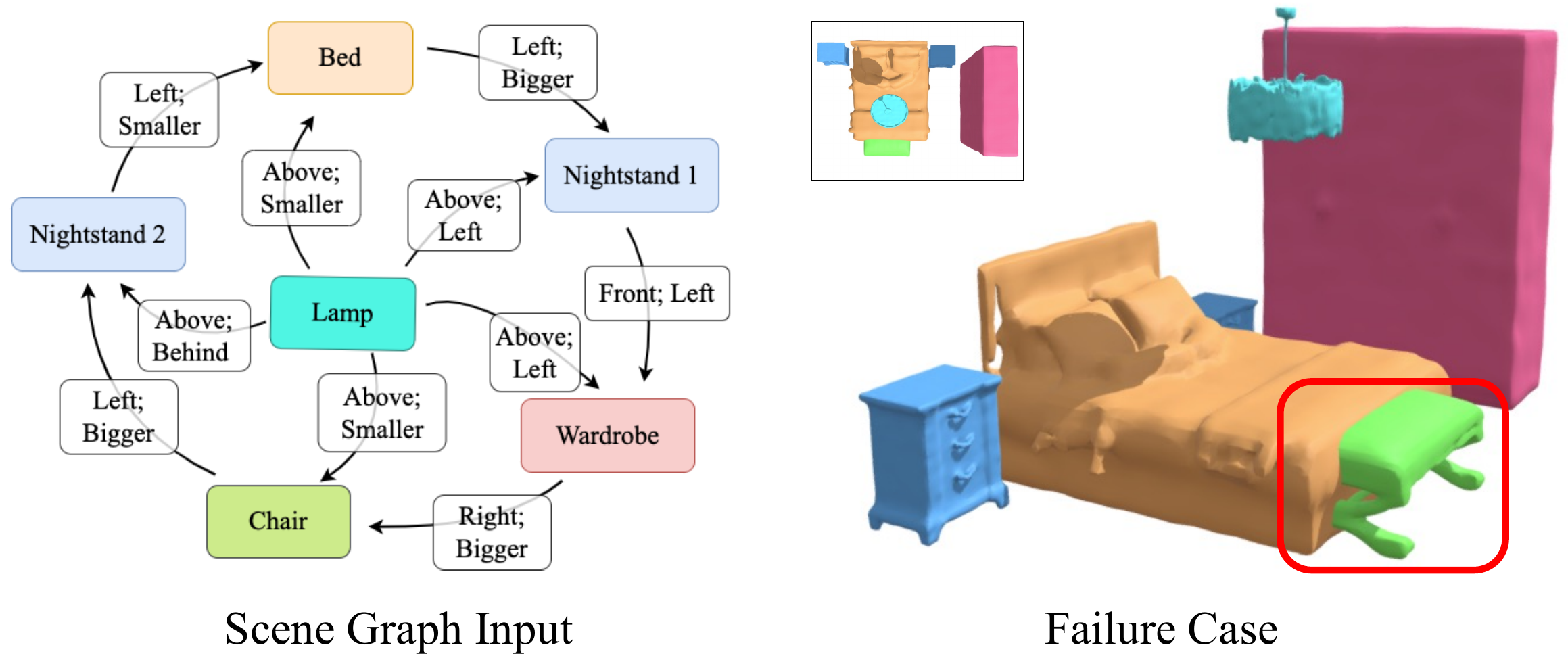}
\captionof{figure}{An interpenetrating  phenomenon.\label{fig:failure}}
\end{minipage}
\begin{minipage}[b]{0.48\textwidth}
\centering
\scalebox{0.85}{
    \begin{tabular}{lccc}
    \toprule
         Condition & FID & KID & mSG \\
     \midrule
         Ours w/o context & 50.24  & 5.63 & 0.59\\
         Ours w/o $E_{r}$ & 51.62  & 5.41 & 0.73 \\
         Ours with concat & 48.57 & 4.25 & 0.71  \\
         \textbf{Ours} & \textbf{45.70} & \textbf{3.84} & \textbf{0.74}\\
     \bottomrule
    \end{tabular}
    }
\vspace{0.5cm}
\captionof{table}{Ablations under three circumstances.\label{tab:abl}}
\end{minipage}
\vspace{-0.4cm}
\end{figure}

We ablate the primary components of CommonScenes in \cref{tab:abl}, including (1) scene generation from the original scene graph without contextual information brought by CLIP features, (2) scene generation from the contextual graph without the participation of the relational encoder $E_{r}$, (3) conditioning with concatenation on diffusion latent code instead of cross-attention, and (4) our final proposed method. We provide the mean FID scores over the scenes, as well as the mean over the hard scene graph constraints (mSG). We observe the benefit of both the context and $E_{r}$ indicated by FID/KID scores, as well as the choice of our conditioning mechanism.
\section{Limitations}
\label{sec:lims}

We address the main aspects of the dataset and the limitations of our method and discuss more in the Supplementary Materials. First, the 3D-FRONT dataset used in our research contains significant noise, which we have mitigated through a post-processing step. Despite this effort, a small proportion of noisy data, specifically interpenetrating furniture instances, remains in the training dataset. 
Consequently, our proposed method and the baseline approaches reflect this during inference, rarely resulting in occurrences of scenes with collided objects. We show some cases in Figure~\ref{fig:failure}. While our method outperforms others by effectively combining shape and layout information, it is essential to note that minor collision issues may still arise.

%% file: tex/qual_comp.tex
\begin{figure}[t!]
    \centering
    \includegraphics[width=1\linewidth]{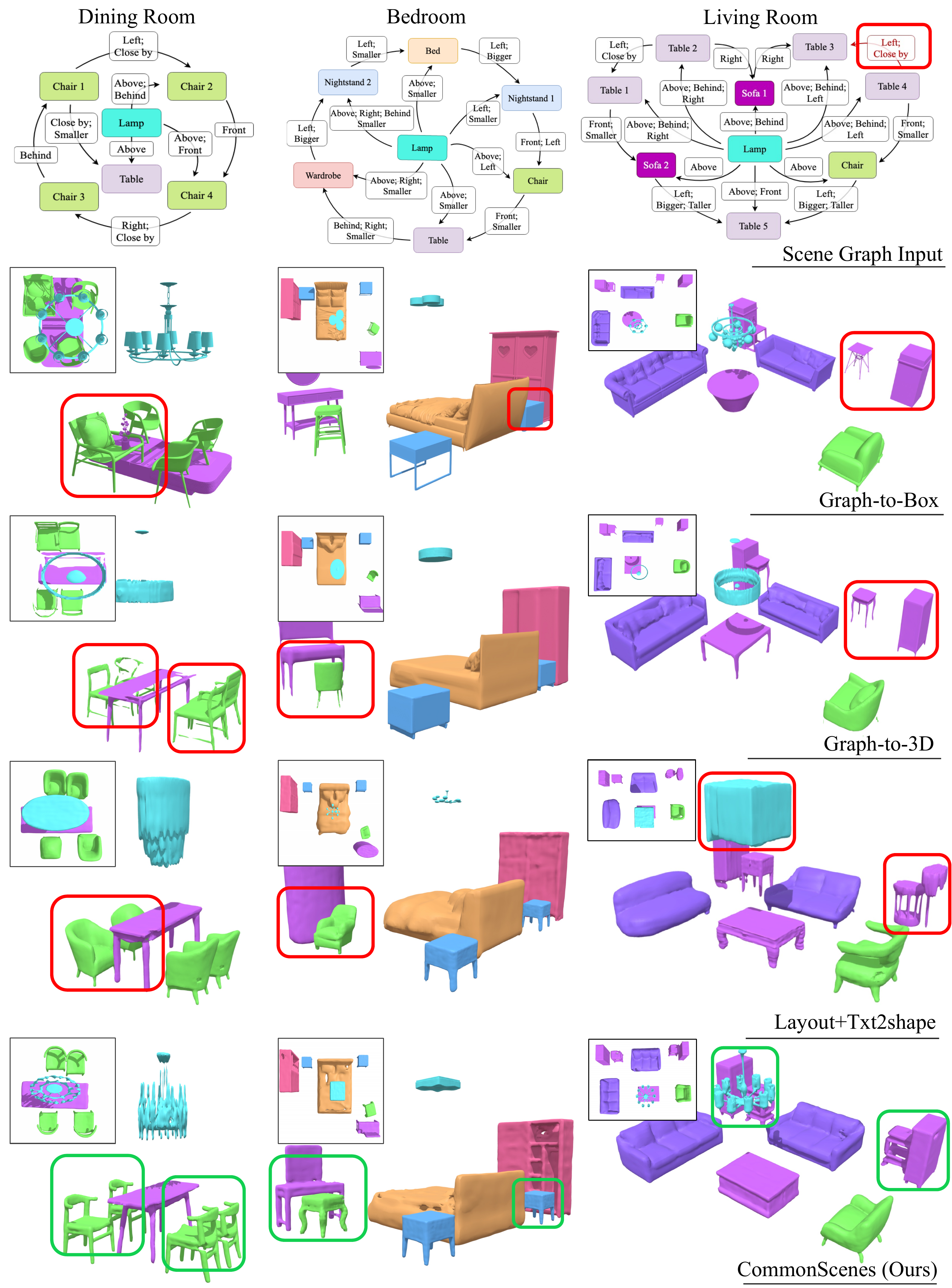}
    \caption{\textbf{Qualitative comparison} The orientations of \texttt{Left/Right} and \texttt{Front/Behind} in the scene graph align with the top-down view. Both scene-object and object-object inconsistencies are highlighted in red rectangles. Green rectangles emphasize the commonsense consistency our method produces.} 
    \label{fig:qual_comp}
    \vspace{-0.3cm}
\end{figure}

%% file: sections/5_Conclusion.tex
\section{Conclusion}
Scene graph-based CSS designs interactive environments suitable for a wide range of usages. Current methods heavily depend on object retrieval or pre-trained shape models, neglecting inter-object relationships and resulting in inconsistent synthesis.
To tackle this problem, we introduce \emph{CommonScenes}, a fully generative model that transforms scene graphs into corresponding commonsense 3D scenes. Our model regresses scene layouts via a variational auto-encoder, while generating satisfying shapes through latent diffusion, gaining higher shape diversity and capturing the global scene-object relationships and local object-object relationships.
Additionally, we annotate a scene graph dataset, \emph{SG-FRONT}, providing object relationships compatible with high-quality object-level meshes. Extensive experiments on \emph{SG-FRONT} show that our method outperforms other methods in terms of generation consistency, quality, and diversity.

%% file: supp.tex
\section*{Supplementary Material of CommonScenes}
In this supplementary, we additionally report the following:
\begin{itemize}
\item Section~\ref{sec:results}: Additional results.
\item Section~\ref{sec:userstudies}: User perceptual studies.
\item Section~\ref{sec:dataset}: SG-FRONT dataset details.
\item Section~\ref{sec:3dssg}: Results on 3DSSG dataset.
\item Section~\ref{sec:qual}: More qualitatives on scene generation.
\item Section~\ref{sec:dis_and_eth}: Discussion and limitations.
\item Section~\ref{sec:training}: Additional training details.
\end{itemize}

We further provide a supplementary video attached to this manuscript to provide spatio-temporal illustrations and further explanations of our method.

\section{Additional Results}
\label{sec:results}

\subsection{Diversity results.}
\begin{figure}[h!]
    \centering
    \includegraphics[width=1.0\linewidth]{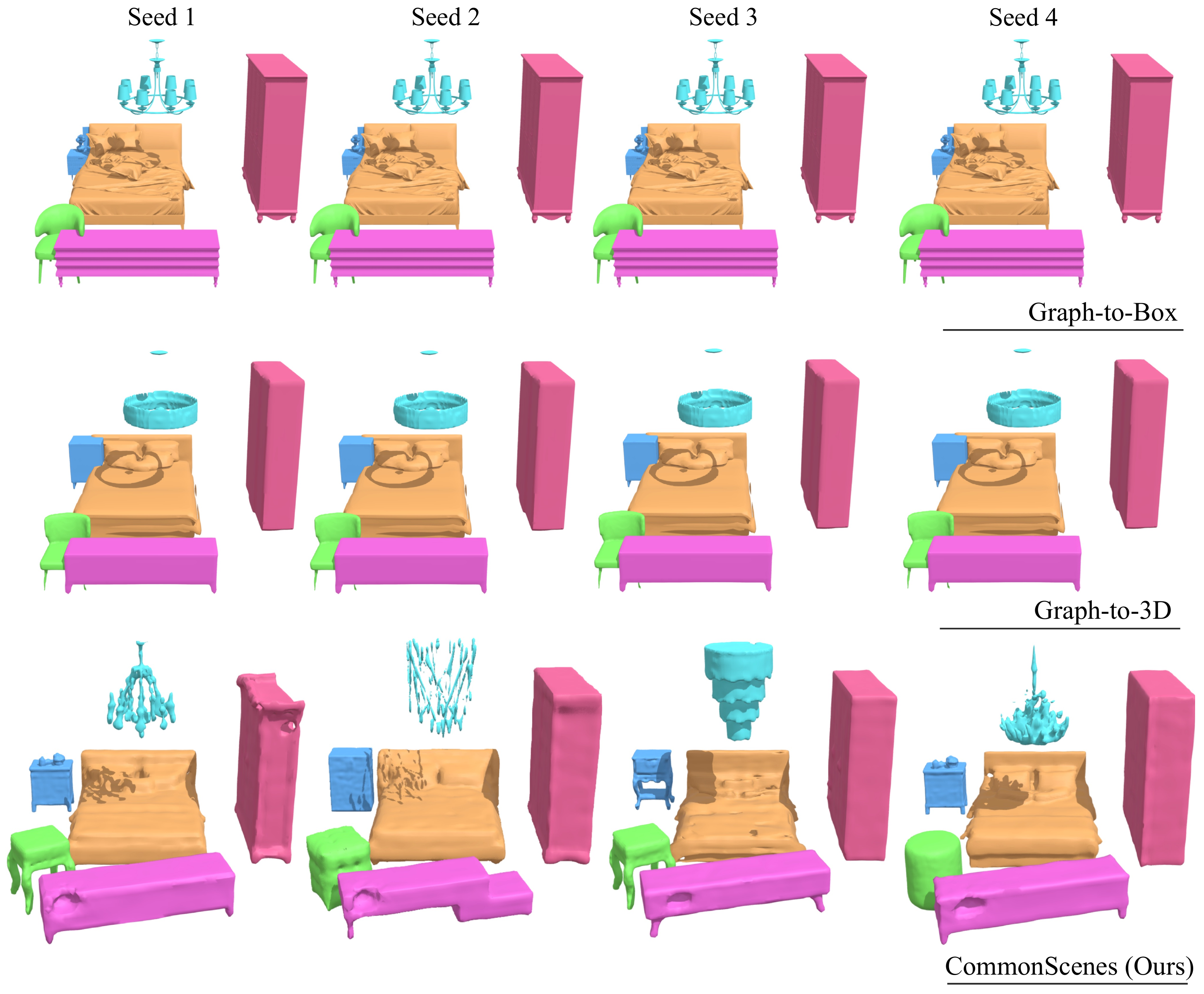}
    \caption{\textbf{Diversity comparison}. Our method shows a huge diversity when generating the same bedroom scene in different rounds.} 
    \label{fig:diversity}
\end{figure}

In Table.~\ref{tab:diver}, we report the results of 10 categories tested on randomly selected 200 test scenes against the state-of-the-art semi-generative method Graph-to-3D~\cite{graph2scene2021} and its corresponding object retrieval method Graph-to-Box. We use Chamfer Distance $(\times 0.001)$ as the metric, following the protocol in Graph-to-3D~\cite{graph2scene2021}. Specifically, we generate each scene 10 times and calculate an average of the chamfer distance between corresponding objects in adjacent scenes. Our method shows a significant improvement upon the diversity leveraging the merits of the diffusion model, with a qualitative example shown in Figure.~\ref{fig:diversity}. The other two methods behave worse because they heavily rely on the shape and embedding of databases, which limits the generative ability.

\begin{table*}[h!]
    \centering
    \scalebox{0.72}{
    \begin{tabular}{l cccccccccc | c}
    \toprule
    Method & Bed & Nightstand & Wardrobe & Chair & Table & Cabinet & Lamp & Shelf & Sofa & TV stand & Total\\
    \midrule 
        Graph-to-Box~\cite{graph2scene2021} & 0.66 & 0.01 & 0.18 & 0.33 & 1.30 & 1.16 & 1.57 & 0.32 & 0.71 & 0.01 & 0.53\\
        Graph-to-3D~\cite{graph2scene2021} & 1.01 & 2.06 & 0.87 & 1.61 & 0.83 & 0.96 & 0.70 & 1.25 & 0.97 & 1.42 & 1.21 \\
        \textbf{Ours}& \textbf{39.59} & \textbf{68.78} & \textbf{20.01} & \textbf{46.03} & \textbf{112.53} & \textbf{32.28} & \textbf{140.56} & \textbf{191.55} & \textbf{58.58} & \textbf{58.08} & \textbf{73.40}\\
    
    \bottomrule
    \end{tabular}
    }
    \caption{\textbf{Diversity performance} on SG-FRONT and 3D-FRONT.}
    \label{tab:diver}
\end{table*}

\subsection{Object-level evaluation.}
Since our objective is scene generation, we use FID/KID as the main metrics for evaluating the scene-level generation quality in the main paper. Although object generation is not the main focus of this work, we believe the object-level analysis is valuable, since this is an integral component of the proposed model. We follow PointFlow~\cite{yang2019pointflow} to report the MMD ($\times0.01$) and COV (\%) for evaluating per-object generation. We collect ground truth objects in each category within the test set. As shown in the first two rows of Tables~\ref{tab:a}, our method shows better performance in both MMD and COV, which highlights the object-level shape generation ability of CommonScenes. 
\begin{table*}[h!]
    \centering
    \scalebox{0.7}{
    \begin{tabular}{l c|cccccccccc}
    \toprule
    Method & Metric & Bed & Nightstand & Wardrobe & Chair & Table & Cabinet & Lamp & Shelf & Sofa & TV stand\\
    \midrule 
        Graph-to-3D~\cite{graph2scene2021} & \multirow{2}{*}{MMD ($\downarrow$)} & 1.56 & 3.91 & 1.66 & 2.68 & 5.77 & 3.67 & 6.53 & 6.66 & 1.30 & 1.08 \\
        \textbf{Ours}& & \textbf{0.49} & \textbf{0.92} & \textbf{0.54} & \textbf{0.99} & \textbf{1.91} & \textbf{0.96} & \textbf{1.50} & \textbf{2.73} & \textbf{0.57} & \textbf{0.29} \\
    \midrule 
        Graph-to-3D~\cite{graph2scene2021} & \multirow{2}{*}{COV ($\%, \uparrow$)} & 4.32 & 1.42 & 5.04 & 6.90 & 6.03 & 3.45 & 2.59 & 13.33 & 0.86 & 1.86 \\
        \textbf{Ours}& & \textbf{24.07} & \textbf{24.17} & \textbf{26.62} & \textbf{26.72} & \textbf{40.52} & \textbf{28.45} & \textbf{36.21} & \textbf{40.00} & \textbf{28.45} & \textbf{33.62} \\
    \midrule 
        Graph-to-3D~\cite{graph2scene2021} & \multirow{2}{*}{1-NNA ($\%, \downarrow)$} & 98.15 & 99.76 & 98.20 & 97.84 & 98.28 & 98.71 & 99.14 & 93.33 & 99.14 & 99.57 \\
        \textbf{Ours}& & \textbf{85.49} & \textbf{95.26} & \textbf{88.13} & \textbf{86.21} & \textbf{75.00} & \textbf{80.17} & \textbf{71.55} & \textbf{66.67} & \textbf{85.34} & \textbf{78.88} \\
    \bottomrule
    \end{tabular}
    }
    \caption{\textbf{Object-level generation performance.} We report MMD($\times0.01$), COV and 1-NNA for evaluating shapes by means of quality and diversity.}
    \label{tab:a}
\end{table*}
We also calculate 1-nearest neighbor accuracy (1-NNA, \%), which directly measures distributional similarity on both diversity and quality. The closer the 1-NNA is to 50\%, the better the shape distribution is captured. It can be observed that our method surpasses Graph-to-3D in the evaluation of distributional similarity. Coupled with the results in Tables 1 and 2, CommonScenes exhibits more plausible object-level generation than the previous state-of-the-art. 
\subsection{Manipulation results.}
\begin{figure}[!ht]
    \centering
    \includegraphics[width=0.8\linewidth]{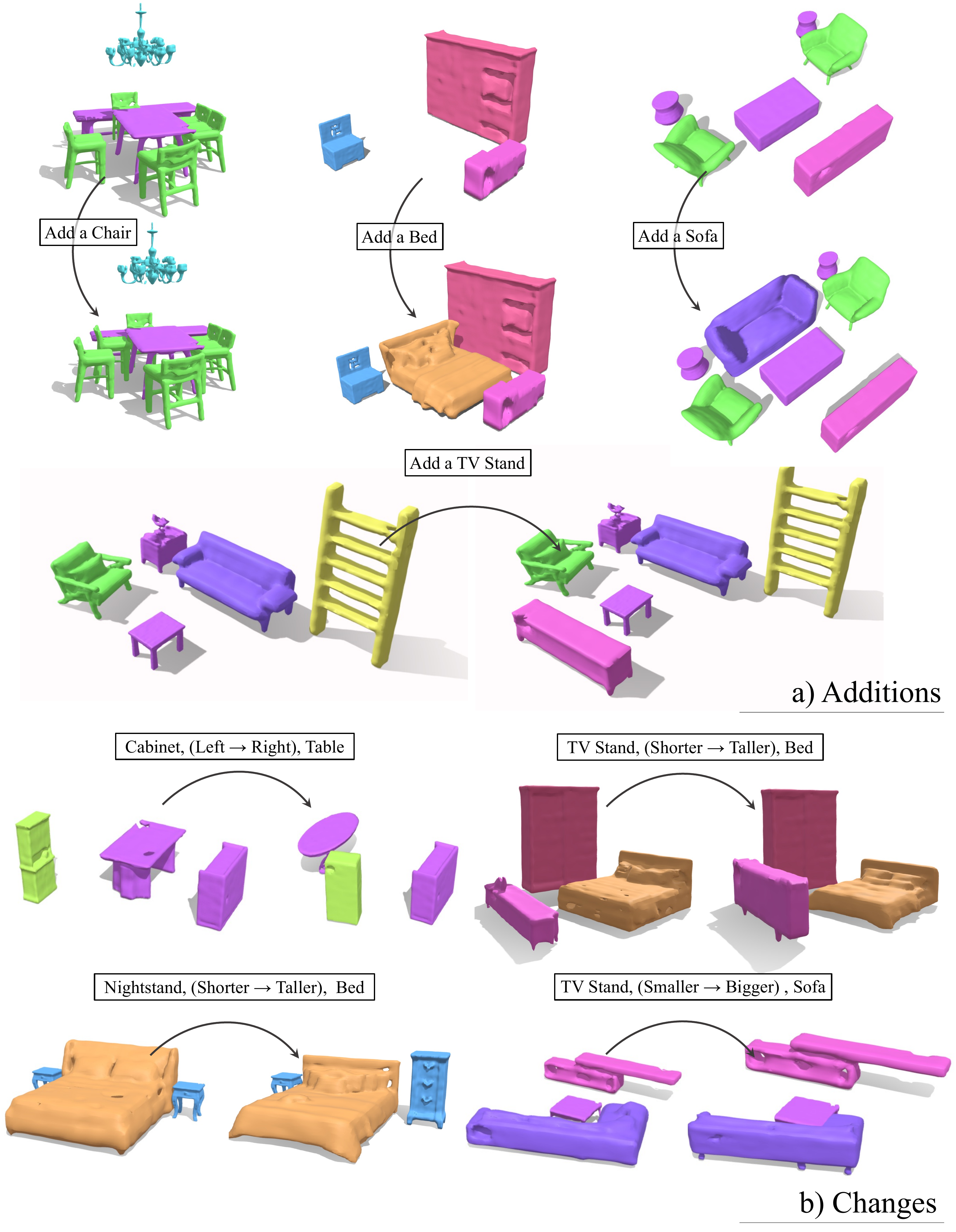}
    \caption{\textbf{Scene manipulation results.} Samples from SG-FRONT and 3D-FRONT depicting object additions and relation changes.} 
    \label{fig:mani}
\end{figure}

Our method inherits the manipulation function from Graph-to-3D, allowing users to manipulate the updated contextual graph during the training and inference phases. As shown in Figure.~\ref{fig:mani}, we provide samples of all three room types (`Bedroom', `Living Room', `Dining Room') for object addition and relation change modes. For instance, in the first scene of Figure~\ref{fig:mani}. a) we first let the model generate six chairs, two tables, and a lamp. Then in the second round, we insert a chair and its edges into other objects in the scene, where the model can still generate the corresponding scenes. In the first scene of Figure.~\ref{fig:mani}. b) the model first generates two tables on the right and one cabinet on the left. In the second round, we manipulate the cabinet to the right side of the left table in the scene. It is an expected sign when the appearances of some objects change, as the second round enjoys different random layout-shape vectors and noise.
Notably, the inserted objects preserve the stylistic consistency within the scene since the sampling is based on the existing contextual knowledge of the scene (\eg, sofa and chair insertions). This is also observed in relation to the change mode, where the generated objects are still realistic after the size or location change. For example, in the bedroom scene where the "nightstand, shorter, bed" triplet is changed to "nightstand, taller, bed", the method can insert plausible objects by replacing the bed with a shorter bed and the nightstand with a taller one, instead of simply stretching and deforming the existing shapes. 

\subsection{Statistical significance test}

We re-run the inference process multiple times with random seeds to test the fluctuation in our scene generation results. Overall, after running the method over ten times, the mean FID and KID scores are in the same range as the results we reported in the main paper Table 1, and the variance in terms of standard deviation on FID is in average $0.05$ scores, where the numbers for bedroom, dining room, and living room are $0.034$, $0.021$, and $0.07$ respectively. This does not alter the comparative results provided in the paper.

\section{User Perceptual Studies}
\label{sec:userstudies}

\begin{figure}[h!]
    \centering
    \includegraphics[width=0.6\linewidth]{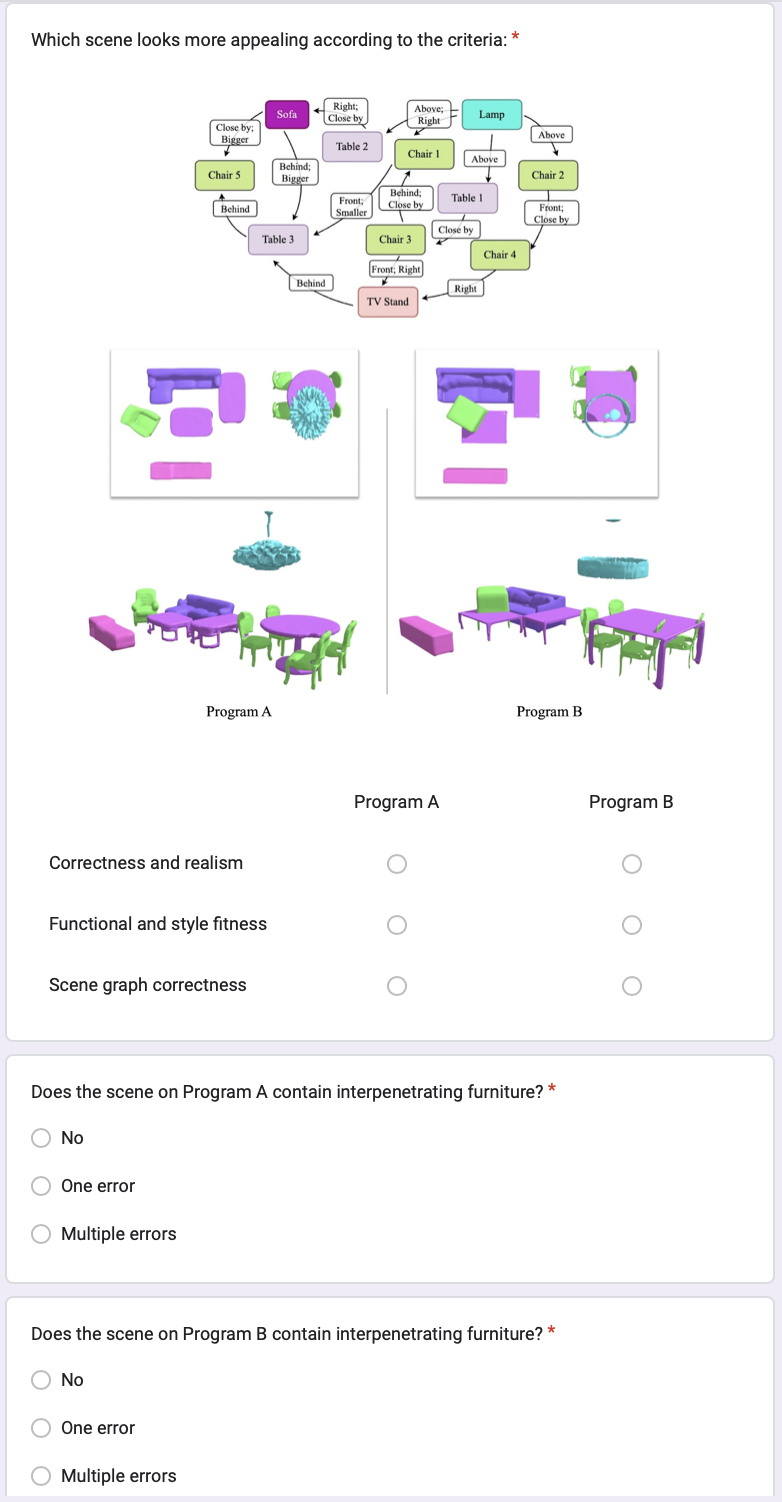}
    \caption{\textbf{User interface} for the perceptual user study.} 
    \label{fig:userstudyss}
\end{figure}

We conducted a perceptual study to evaluate the quality of our generated scenes. 
We generated scenes from Graph-to-3D, \textit{} and our method for this. At every step, we provide samples from paired methods and ask users to select the better one according to the criteria, (1) Global correctness and realism (\textit{Does the arrangement of objects look correct?}), (2) functional and style fitness (\textit{Does the scene look stylistically coherent?}), (3) scene graph correctness. Additionally, we asked for the number of errors in interpenetrating furniture for each scene to evaluate the visible errors. Three answers were shown (1) No errors, (2) One error, and (3) Multiple errors to prevent attention degradation of users. We randomly sample $\sim$ 20 scenes covering all scene types. For each scene, we provide a top-down view rendering and a side-view rendering to ensure the visibility of the entire scene. We illustrate the user interface of this study \cref{fig:userstudyss}.

\begin{figure}[h!]
    \centering
    \includegraphics[width=0.32\linewidth]{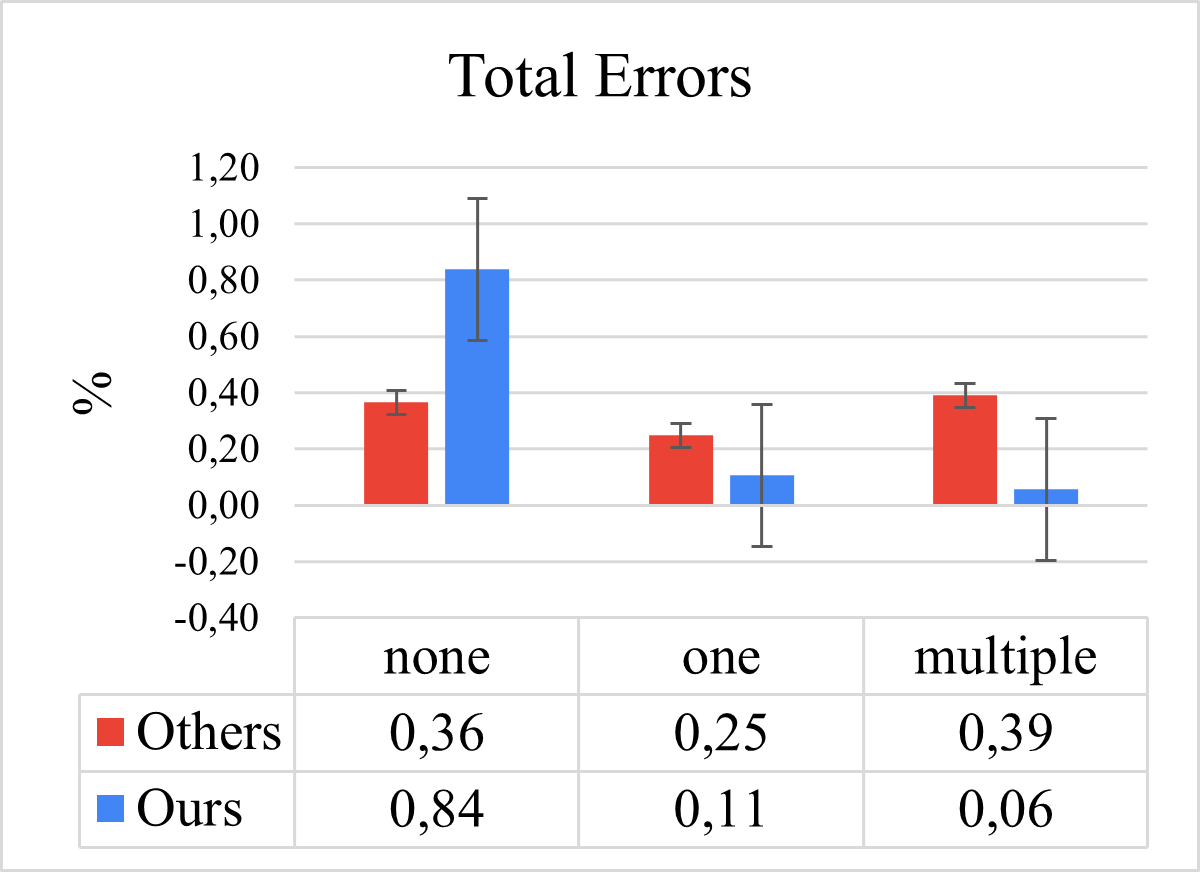}
    \includegraphics[width=0.32\linewidth]{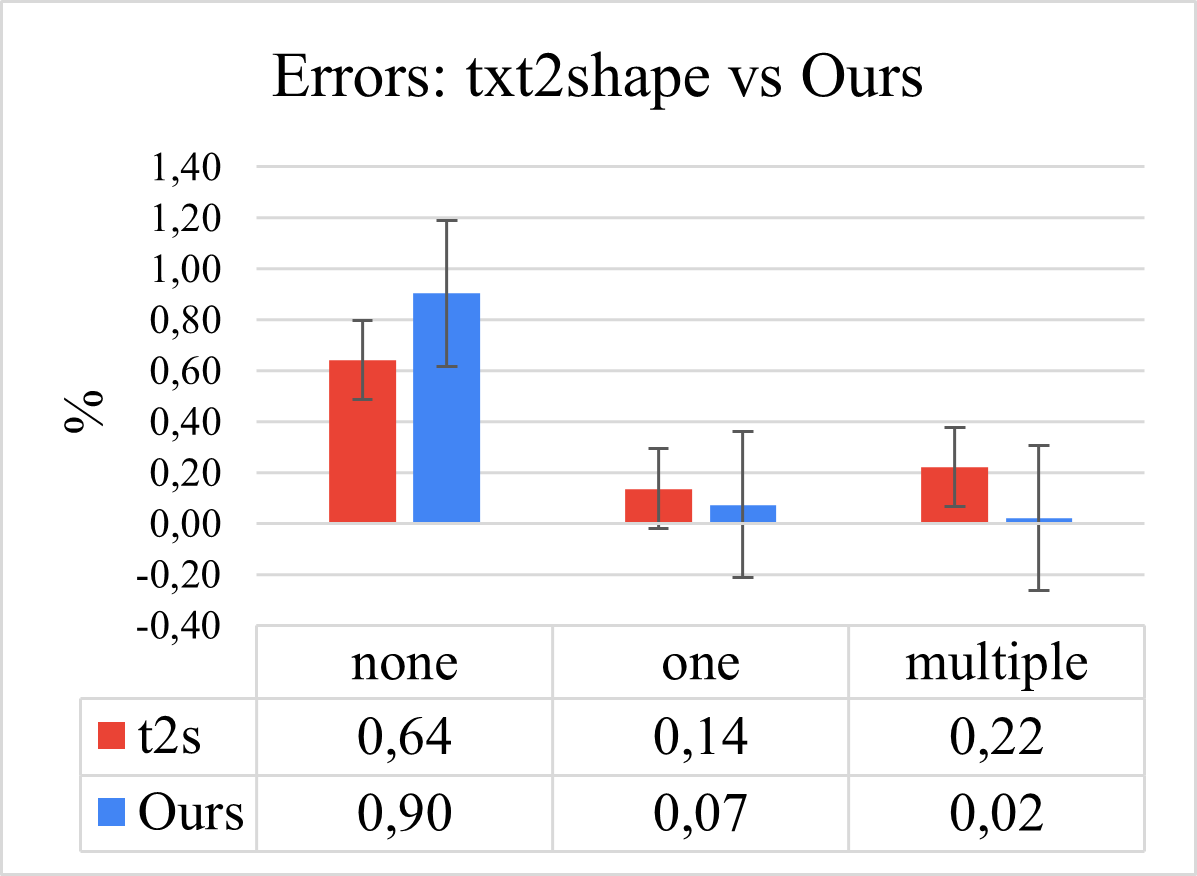}
    \includegraphics[width=0.32\linewidth]{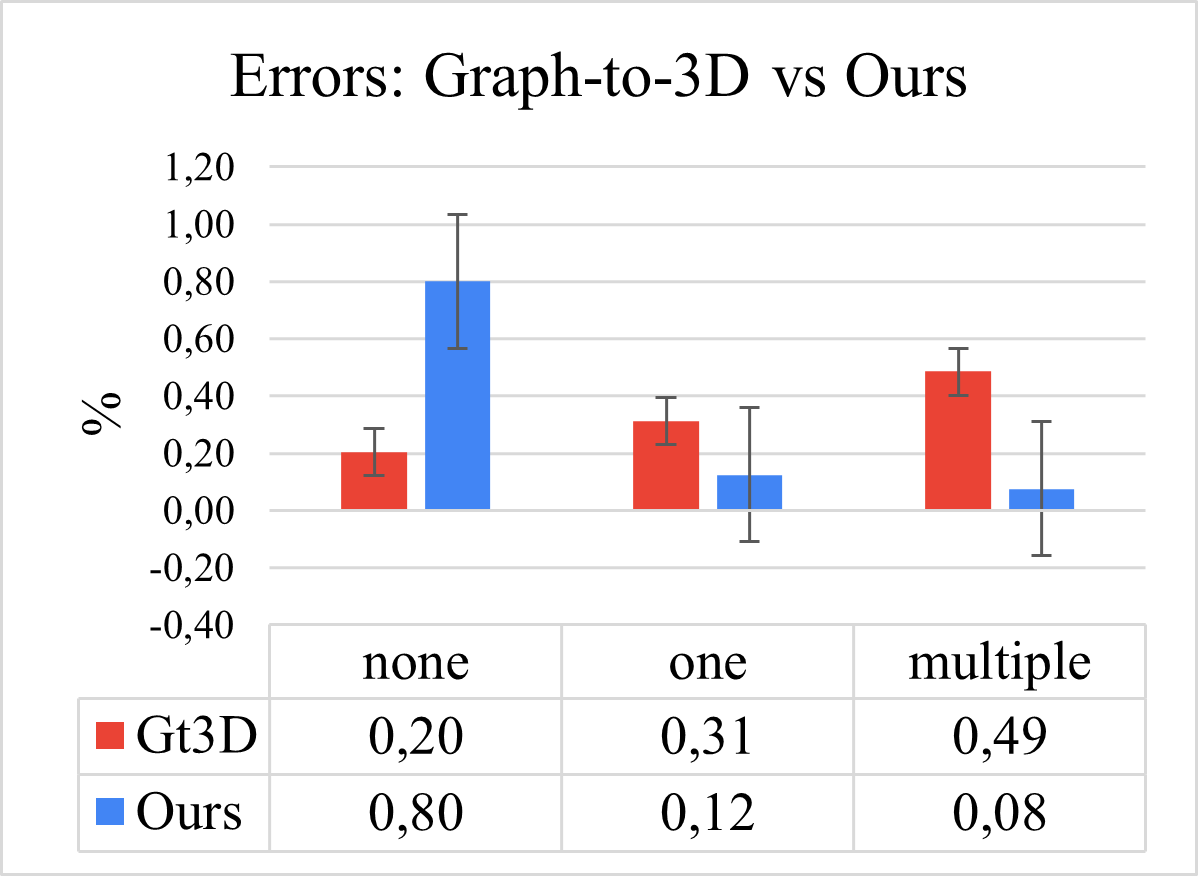}
    \caption{\textbf{Error bars} to indicate the number of errors in scenes observed by the user study participants.} 
    \label{fig:errorbar}
\end{figure}

Overall, $\sim25$ subjects from various nationalities, professional backgrounds, and ages participated in our study. We provide the outcomes of this study in \cref{tab:percept} and \cref{fig:errorbar}. The scenes generated from our method were preferred over the other methods $83\%$ for correctness and realism, $80\%$ for style fitness, and $87\%$ for correctness. Where layout+txt2shape was instead preferred $16\%$, $18\%$ and $6\%$ and Graph-to-3D $17\%$, $20\%$ and $15\%$, respectively. Interestingly, our method was preferred over the scene graph correctness much higher than other aspects, whereas layout+txt2shape was the lowest, proving the necessity of information sharing between layout and 3D shapes. Unsurprisingly, Graph-to-3D was preferred more than layout+txt2shape because of the same reason. The inclination towards our method was the highest in all aspects, indicating that our proposed full generative method could generate commonsense scenes. 

Additionally, we can observe the error comparison in \cref{fig:errorbar} that the users indicated $84\%$ scenes having no error, $11\%$ one error, and $6\%$ as more than one error, where all others had $36\%$ none, $25\%$ one, and $39\%$ multiple errors. We also provide the statistics separately for each method. Here, it can be seen that on the scenes compared with the layout+txt2shape method, ours had $90\%$ no errors and $2\%$ multiple errors, whereas the compared method had $64\%$ none and $22\%$ multiple. In the scenes that were compared across Graph-to-3D and ours, they had $20\%$ no errors and $49\%$ multiple errors, whereas our method $80\%$ none and only $8\%$ multiple errors. This result shows that our method improves the error of interpenetrating objects compared to the other baselines.

\begin{table*}[h!]
    \centering
    \scalebox{0.80}{
    \begin{tabular}{lc|ccc}
    \toprule
         Method &  Condition & Corr. \& Real. & Style Fit. & Scene graph \\
     \midrule
         Layout+txt2shape & Ours & 0.16 & 0.18 & 0.06 \\
         Graph-to-3D~\cite{graph2scene2021} & Ours & 0.17 & 0.20 & 0.15 \\
         Ours & All & 0.83 & 0.80 & 0.87 \\
     \bottomrule
    \end{tabular}
    }
    \caption{\textbf{Perceptual study results.} Results over paired study, where the users indicated the preference over A/B scene comparison.}
    \label{tab:percept}
\end{table*}

\section{SG-FRONT Dataset Details}
\label{sec:dataset}

\begin{figure}[h]

    \begin{subfigure}{\linewidth}
        \centering
    \includegraphics[width=0.49\linewidth]{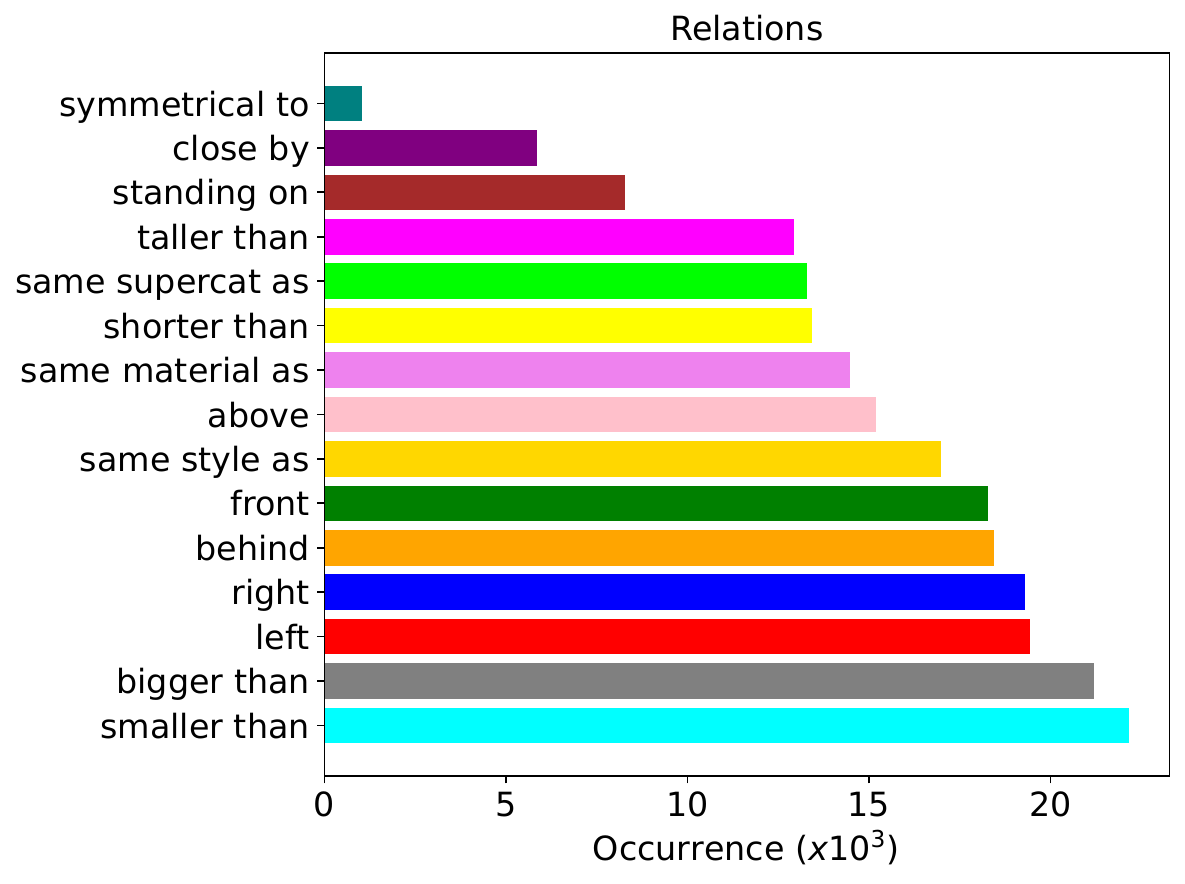}
    \includegraphics[width=0.49\linewidth]{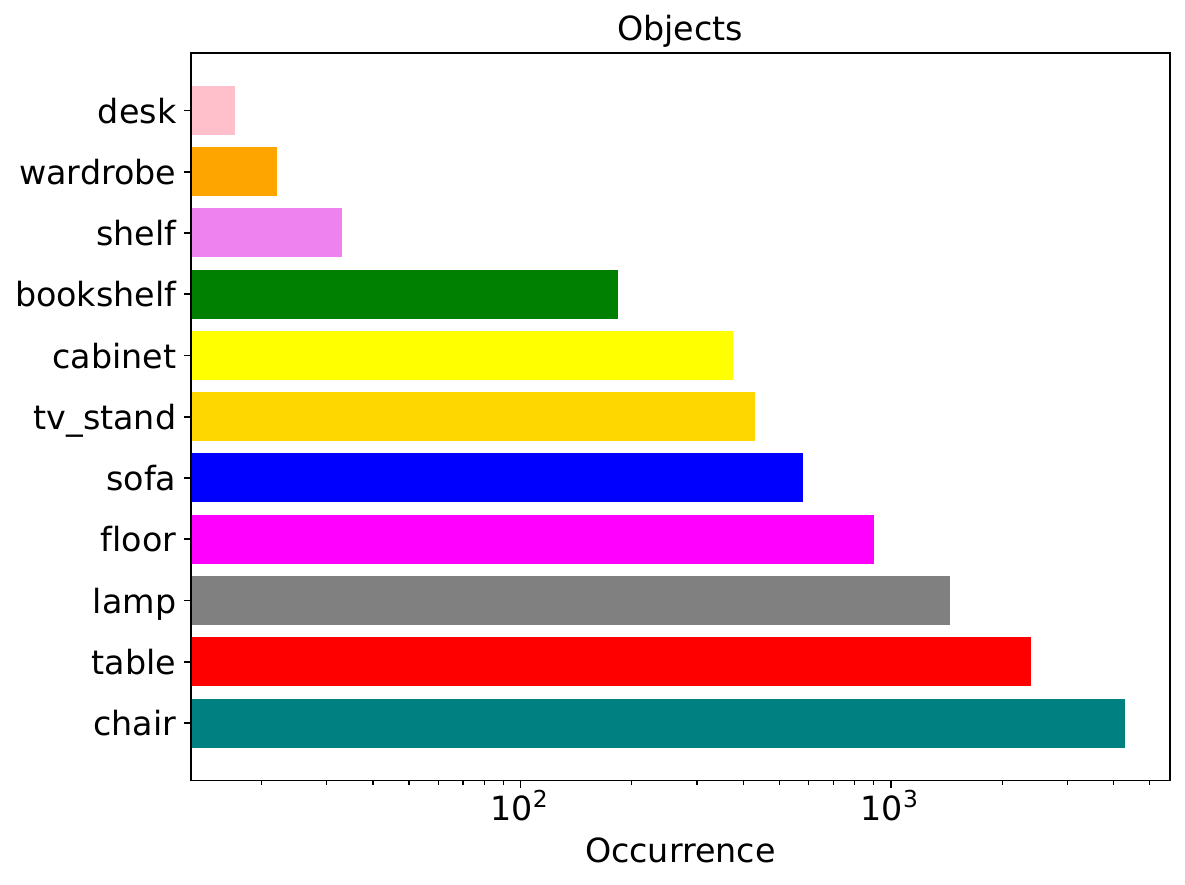}
    \caption{Dining room relationships.} 
    \end{subfigure}
    
    \hfill

    \begin{subfigure}{\linewidth}
        \centering
    \includegraphics[width=0.49\linewidth]{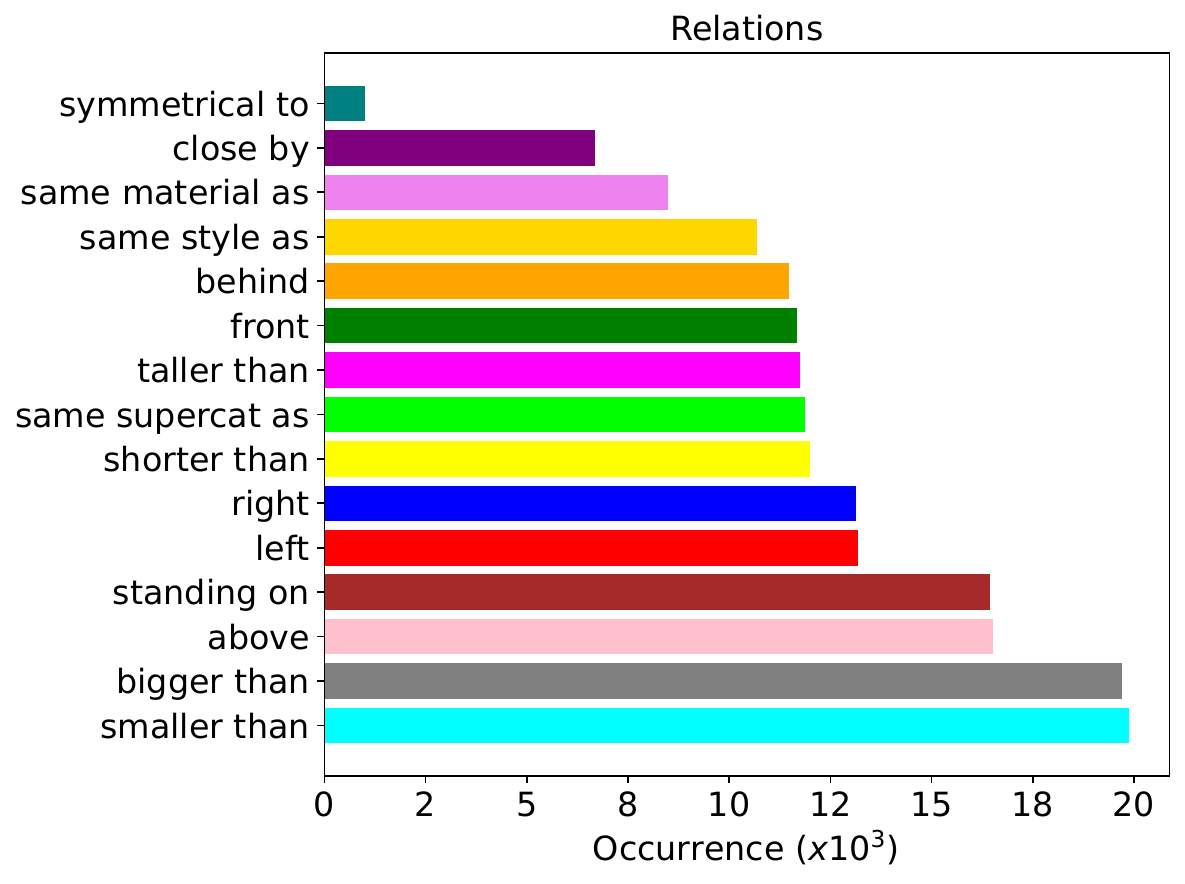}
    \includegraphics[width=0.49\linewidth]{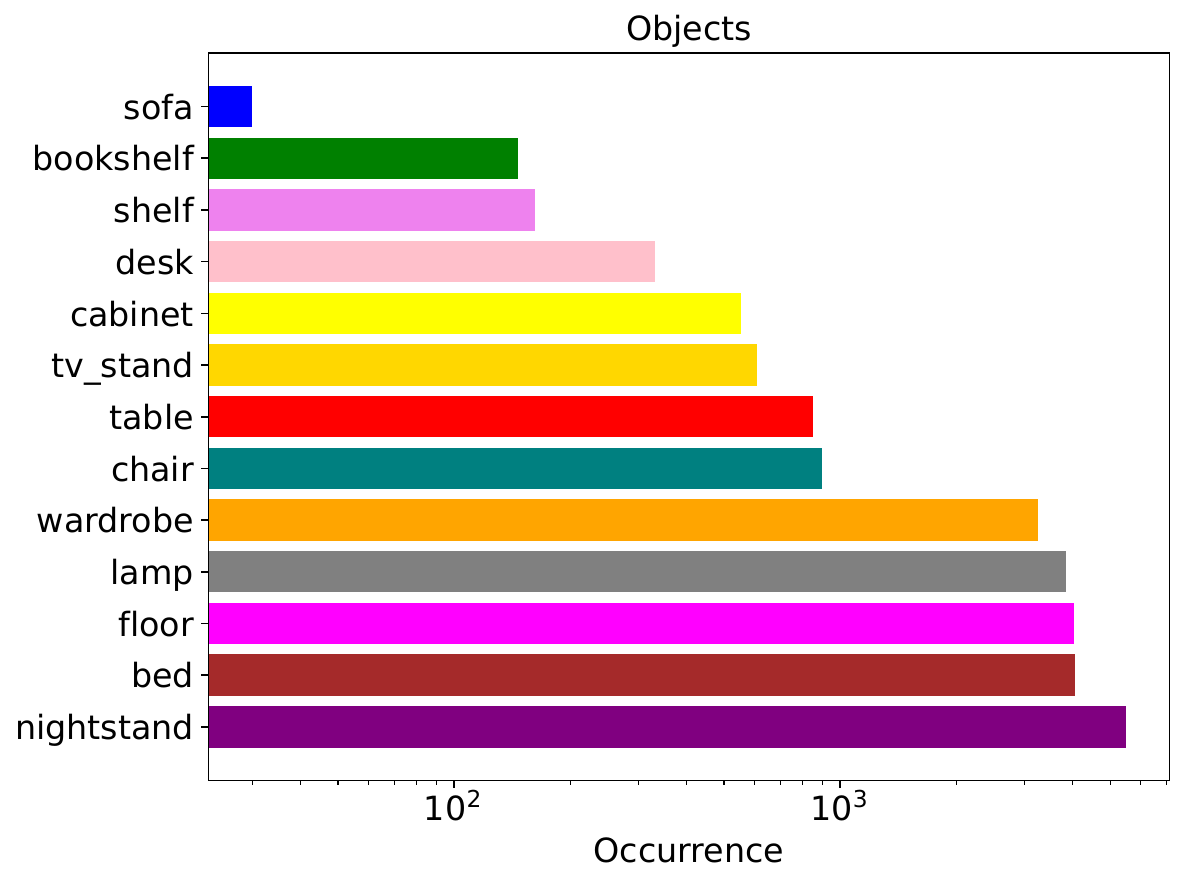}
    \caption{Bedroom relationships.} 
    \end{subfigure}

    \hfill

    \begin{subfigure}{\linewidth}
        \centering
    \includegraphics[width=0.49\linewidth]{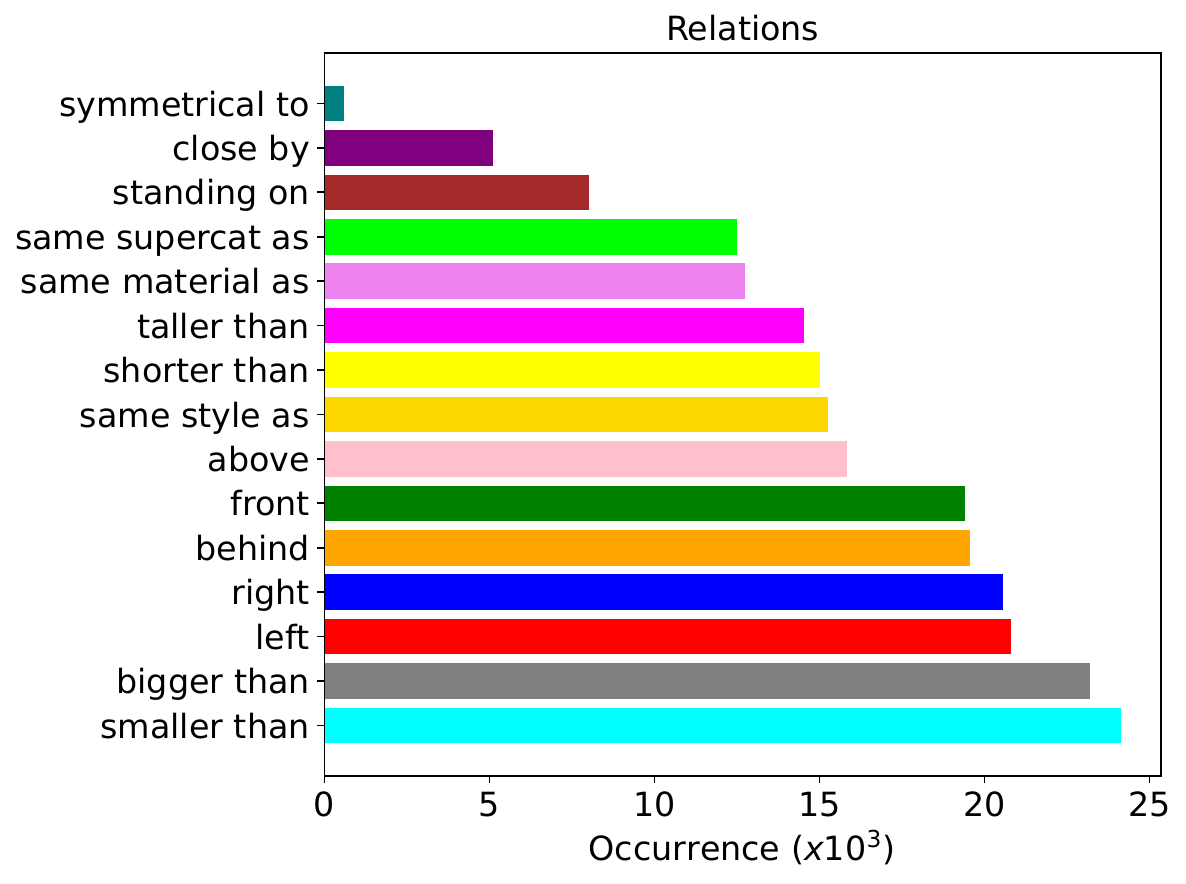}
    \includegraphics[width=0.49\linewidth]{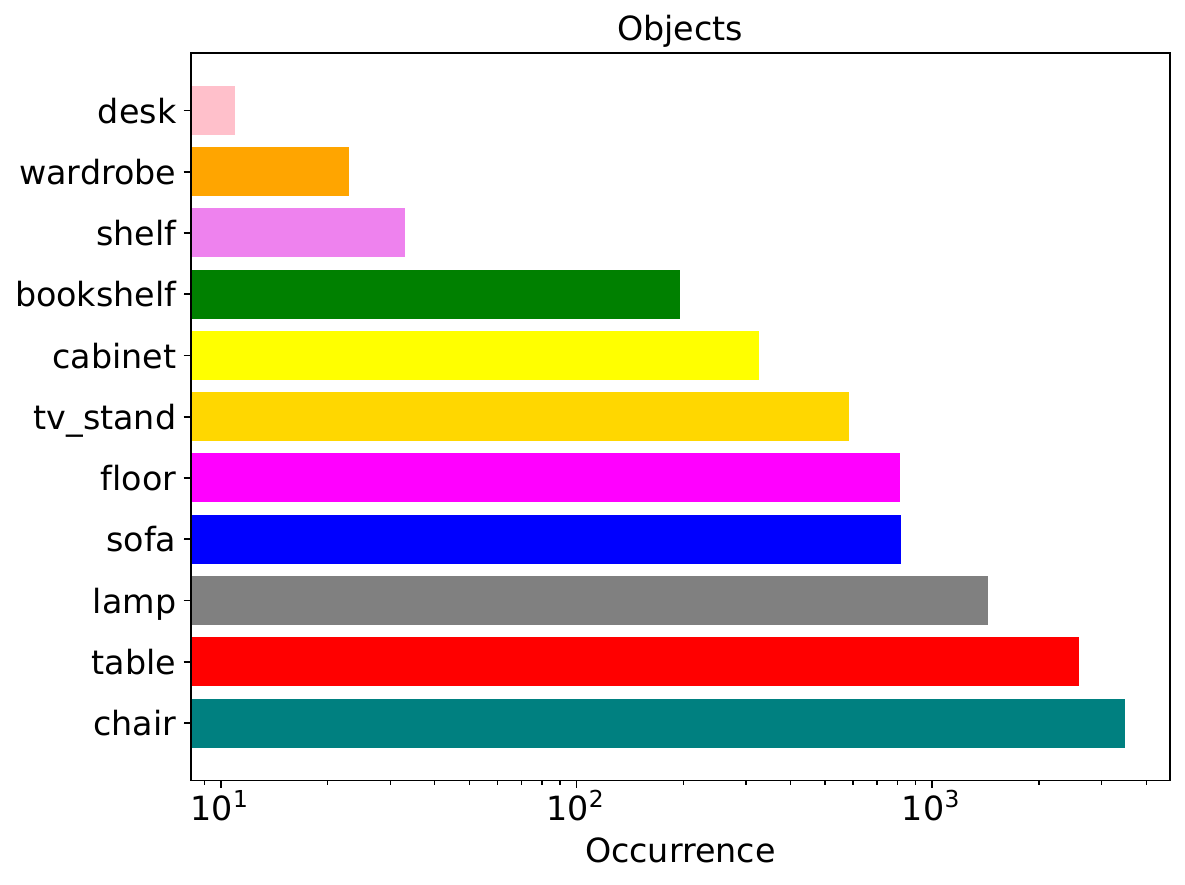}
    \caption{Living room relationships.} 
    \end{subfigure}
    
    \caption{\textbf{SG-FRONT dataset statistics} for (a) Dining room, (b) bedroom, and (c) Living room scenes. Relationships are provided on the left, and the object occurrences on the right.}
    \label{fig:dataset}
\end{figure}

We have annotated the 3D-FRONT dataset~\cite{3dfront} with scene graphs to create the SG-FRONT dataset. For annotating the layouts with scene graphs, we follow the previous works 3DSSG \cite{3DSSG2020} and 4D-OR \cite{4D_OR} with a semi-automatic annotation approach. The dataset statistics regarding the number of relations and objects per room type are summarized in Figure.~\ref{fig:dataset}. 

Furthermore, the original 3D-FRONT dataset comprises a large variety of scenes, including noise and unrealistic clutter, and previous work used a preprocessed version of this dataset that encompasses three room types (living room, dining room, and bedroom) and removes extremely cluttered scenes, similar scenes, or the scenes that contain a high amount of collision. We follow the same preprocessing criteria and train/test splits provided at ATISS \cite{Paschalidou2021NEURIPS}. The whole SG-FRONT contains the scene graph annotations of 4,041 bedrooms, 900 dining rooms, and 813 living rooms, containing 5,754 scenes in total, the same rooms as processed 3D-FRONT. There are 4,233 distinct objects with 45K instance labels in the scenes.

\section{Results on 3DSSG Dataset}
\label{sec:3dssg}

\paragraph{Discussion over 3RScan versus 3D-FRONT.}
The existing scene graph datasets lack high-quality object-level meshes to facilitate our task. One previously introduced such dataset, 3DSSG, is built upon the 3RScan dataset \cite{Wald2019RIO}, which consisted of scanned indoor environments from a Tango mobile phone to detect camera relocalization in changing indoor environments. Further, the semantic scene graph annotation, 3DSSG \cite{3DSSG2020}, is provided means to alleviate the problem of camera relocalization from 3D scene graphs. Graph-to-3D built their method with this dataset since it was the only available indoor 3D scene graph dataset at the time. Their AtlasNet version could generate scenes in sparse point cloud format, yet, it is inherently inferior to the DeepSDF version in terms of the quality of dense reconstruction. However, since 3RScan cannot be used to train the DeepSDF branch due to its noisiness and incomplete meshes, Graph-to-3D trains on object-level meshes from ShapeNet \cite{chang2015shapenet}. Although it can successfully enable the method to proceed, the fragments between 3RScan and ShapeNet prohibit the method from being fully functional and tested correctly. To achieve high-quality scene generation and a fair baseline comparison, we construct a scene graph dataset SG-FRONT based on 3D-FRONT \cite{3dfront}. 3D-FRONT is a large-scale indoor synthetic dataset with professionally designed layouts and a large variety of objects reflecting natural environments. The scenes contain stylistically consistent 3D objects placed according to the choice of interior designers. Compared to 3DSSG, this dataset facilitates 3D scene generation research by providing high-quality watertight 3D object meshes from 3D-FUTURE~\cite{fu20213d}.

\paragraph{Comparison to baselines.} For complementary reasons, we provide results on the 3DSSG dataset as well. Since the usage of 3DSSG is not directly possible \cite{graph2scene2021} and would give suboptimal results within our method with the ShapeNet dataset, we evaluate our method and compare it with the previous methods in terms of layout generation. We again use scene graph constraints to measure layouts. In this aspect, we use the original five metrics (\texttt{left/right}, \texttt{front/behind}, \texttt{smaller/larger}, \texttt{lower/higher}, and \texttt{same as}), which was excluded in previous work \cite{graph2scene2021}. Scene generation results are shown in \cref{tab:generation_3dssg}, and the scene manipulation results (in addition and change modes) are given in \cref{tab:mani_3dssf}. In the generation phase, our method has impressive results compared with others, showing that our contextual graph can improve the performance of layout understanding. In manipulation, our method still shows strong results, where \texttt{left/right}, \texttt{smaller/larger}, \texttt{lower/higher}, and \texttt{same as} are better than others.

\begin{table*}[h!]
    \centering
    \scalebox{0.8}{
    \begin{tabular}{l|ccccc}
    \midrule 
        \multirow{2}{*}{Method} & left / & front / & smaller / & lower / & \multirow{2}{*}{same as}    \\
         & right & behind & larger & higher & \\
    \midrule 
        3D-SLN \cite{Luo_2020_CVPR} & 0.74 & 0.69 & 0.77 & 0.85  & \textbf{1.00} \\
        Graph-to-Box & 0.82 & 0.78 & 0.90 & \textbf{0.95}  & \textbf{1.00} \\
    \midrule 
    \textbf{Ours} w/o SB & \textbf{0.90} & \textbf{0.84} & \textbf{0.98} & \textbf{0.95} &  \textbf{1.00} \\
    \midrule
    \end{tabular}
    }
    \caption{\textbf{Scene generation results on 3DSSG} in terms of graph constraints (higher is better).}
    \label{tab:generation_3dssg}
\end{table*}

\begin{table*}[h!]
    \centering
    \scalebox{0.85}{
    \begin{tabular}{l c cccc c}
    \toprule
    \multirow{2}{*}{Method} &\multirow{2}{*}{Mode}& left / & front / & smaller / & lower / & \multirow{2}{*}{same as}   \\
     &   & right & behind & larger & higher & \\
    \midrule 
        3D-SLN~\cite{Luo_2020_CVPR} & \multirow{4}{*}{change} & 0.62 & 0.62 & 0.66 & 0.67 & 0.99   \\
        Graph-to-Box~\cite{graph2scene2021} & & 0.65 & 0.66 & 0.73 & 0.74 & 0.98\\
        \cmidrule{1-1} \cmidrule{3-7}
        \textbf{Ours} w/o SB & & \textbf{0.71} & \textbf{0.71} & \textbf{0.76} & \textbf{0.80} & \textbf{1.00}  \\
        
        \midrule \midrule
        3D-SLN~ \cite{Luo_2020_CVPR} & \multirow{4}{*}{addition} & 0.62 & \textbf{0.63} & 0.78 & 0.76 & 0.91    \\
        Graph-to-Box~\cite{graph2scene2021} &  & 0.63 & 0.61 & 0.93 & 0.80 & 0.86 \\
        \cmidrule{1-1} \cmidrule{3-7}
        \textbf{Ours} w/o SB & & \textbf{0.72} & 0.62 & \textbf{0.94} & \textbf{0.90} & \textbf{1.00}  \\

    \midrule
    \end{tabular}
    }
    \caption{\textbf{Scene manipulation results on 3DSSG} in terms of graph constraints (higher is better). Top: Relationship change mode. Bottom: Node addition mode.}
    \label{tab:mani_3dssf}
\end{table*}

\section{More Qualitatives on Scene Generation}
\label{sec:qual}
\begin{figure}
    \centering
    \includegraphics[width=1.0\linewidth]{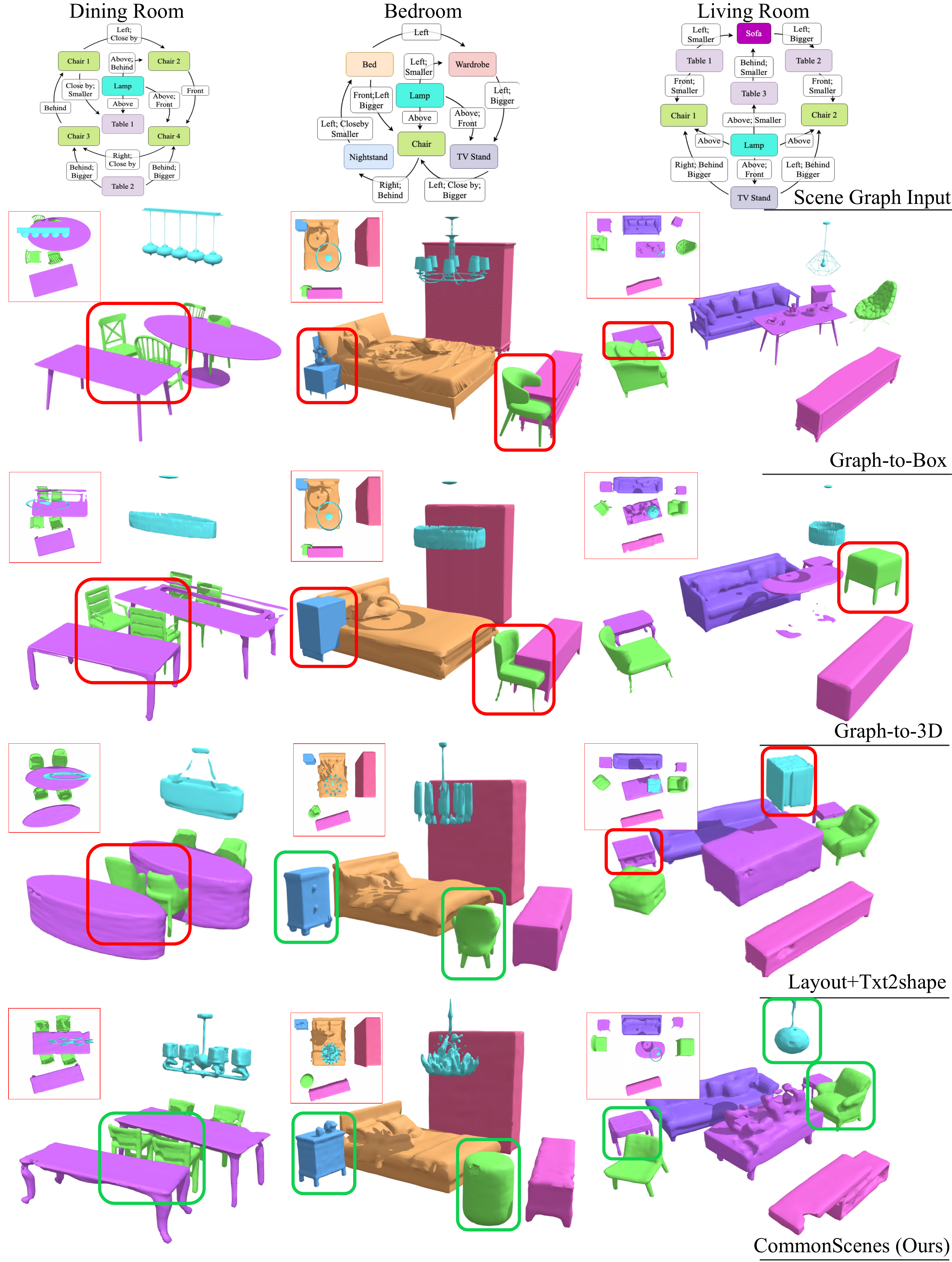}
    \caption{\textbf{Additional generation results.} Red rectangles show both the scene-object and object-object inconsistency, while green ones highlight the reasonable and commonsense settings.} 
    \label{fig:gen3}
\end{figure}

\begin{figure}
    \centering
    \includegraphics[width=1.0\linewidth]{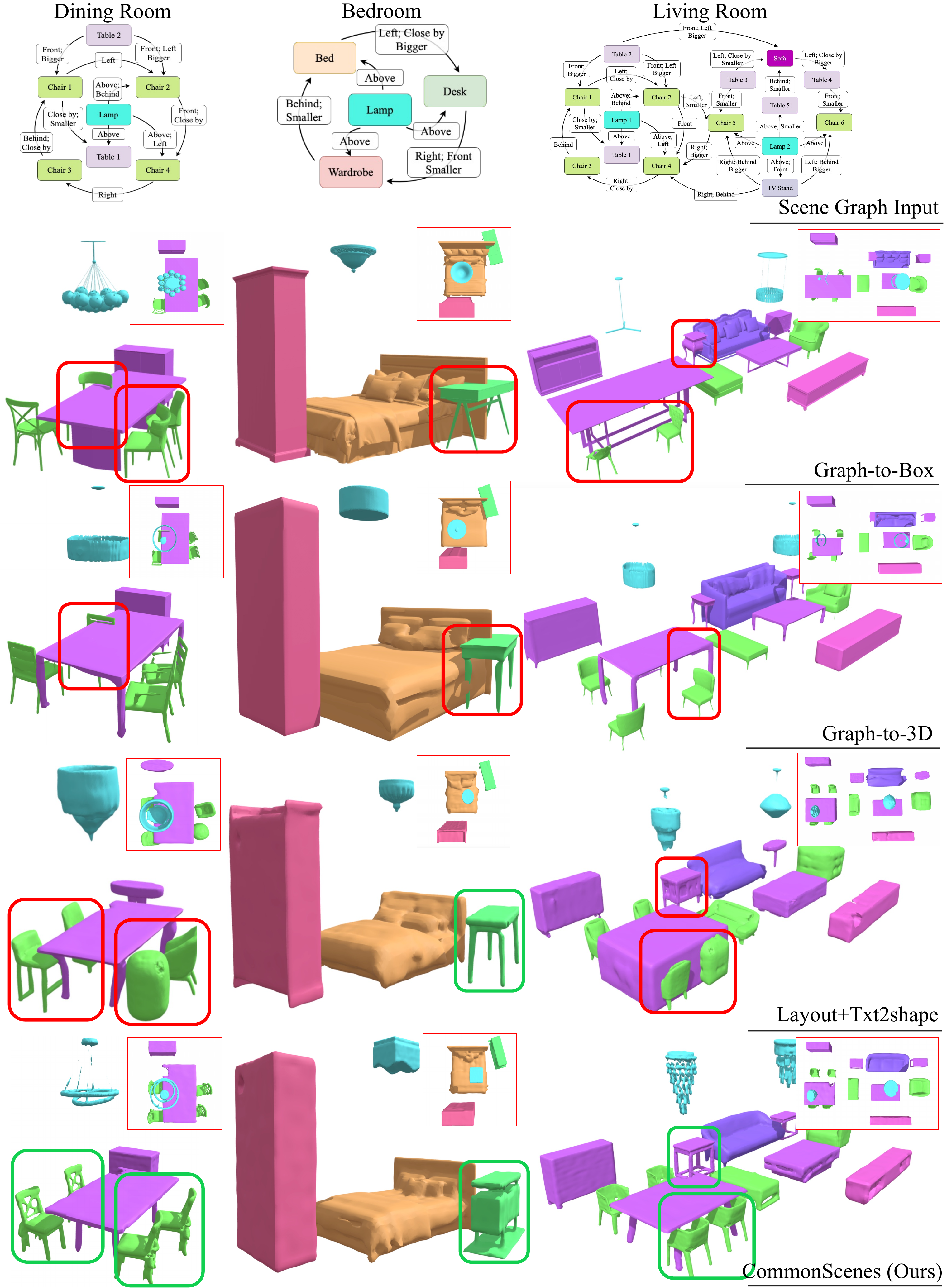}
    \caption{\textbf{Additional generation results.} Red rectangles show both the scene-object and object-object inconsistency, while green ones highlight the reasonable and commonsense settings.} 
    \label{fig:gen4}
\end{figure}

We show more quantitative results in Figure.~\ref{fig:gen3}, and~\ref{fig:gen4} to illustrate that our method can achieve realistic generation, higher object-object and scene-object consistency. Generated shapes should be realistic, while Layout+txt2shape can only randomly generate a lamp but cannot consider whether the real size matches the bounding box size in Fig.~\ref{fig:gen3}, making it stretch too much to be unrealistic. In contrast to these methods, we can achieve every requirement in the scene. For the example of object-object consistency in Figure.~\ref{fig:gen3}, in the dining room, the other three methods cannot generate a suit of dining chairs, while our method can achieve the goal. For the scene-object consistency, in the living room in Fig.~\ref{fig:gen4}, even though Graph-to-3D can generate consistent chairs, but they are not suitable with the table, e.g., incompatible height and unaligned orientation.

\section{Discussion}
\label{sec:dis_and_eth}

Besides the main limitation mentioned in the main paper, we have deliberately excluded texture and material information from our methodology and focused on generating stylistically coherent and controllable 3D scenes. Incorporating additional texture/material details would introduce a new dimension of complexity to the method, as modeling unbounded 3D shapes with materials is not straightforward. Hence, including texture and material information is an exciting avenue for future research.

Regarding ethical considerations, our method does not involve using human or animal subjects, nor does it introduce direct ethical implications. However, being a generative model, it shares ethical concerns commonly associated with other generative models, particularly regarding potential misuse. We are hopeful that the broad impact of generative models will outweigh any negative use cases and that the wider community can leverage this powerful technology for positive advancements.

\section{Additional Training Details}
\label{sec:training}

\subsection{Implementation Details.}

\textbf{Trainval and test splits.} We train and test all models on SG-FRONT and 3D-FRONT, containing 4,041 bedrooms, 900 dining rooms, and 813 living rooms. The training split contains 3,879 bedrooms, 614 dining rooms, and 544 living rooms, with the rest as the test split.

\begin{figure}[h!]
    \centering
    \includegraphics[width=1.0\linewidth]{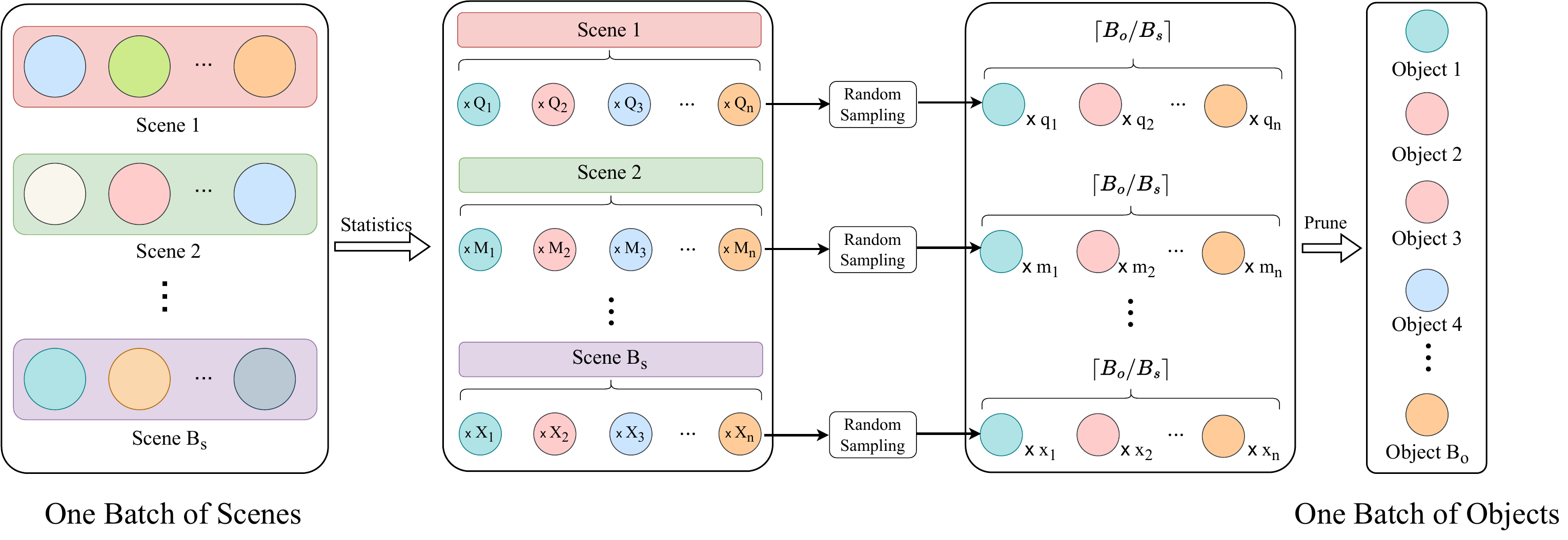}
    \caption{\textbf{Uniformed Sampling.} Objects are represented as colored circles, while scenes are colored rectangles.} 
    \label{fig:batch}
\end{figure}

\textbf{Batch sizes.} The two branches use individual batches in terms of the different training objectives. The layout branch uses a scene batch during one training step, containing all bounding boxes in $B_s$ scenes. The shape branch is supposed to take an equal amount of the relation embeddings out of the relation encoder $E_r$. However, this way is prohibited by the limitation of the memory of the GPU. The shape branch instead takes an embedding batch containing $B_o$ embeddings of the counterpart objects sampled by our \textit{Uniformed Sampling} strategy shown in Figure~\ref{fig:batch}. Our training strategy can support sufficient training using little memory storage and data balance. Given all objects (colored balls) in $B_s$ scenes (colored rectangles), we first set the sampling goal as to obtain $\lceil B_o / B_s \rceil$ objects in each scene and collect all objects in each class of $n$ classes in the scene. Take Scene 1 as an example. We divide all $Q$ objects into $n$ classes: $Q=\sum_{i=1}^n Q_i, Q_i \geq 0$, where $Q_i$ means the number of objects in the $i$-th class, and similarly treat other scenes. We can achieve class-aware sampling in this way to ensure that every class in each scene experiences the training process. Then we sample an object set $\{q_i|q_i \geq 0, i=1, \ldots, n\}$ out of $Q$ objects using random sampling, where $\lceil B_o / B_s \rceil = \sum_{i=0}^{n}q_i, i=1, \ldots, n$. We combine the sampling sets of all scenes and prune objects to $B_o$. Finally, we feed the corresponding relation embeddings into the shape branch. We set $B_s$ as 10 and $B_o$ as 40 in the experiment.

\subsection{Baseline training.}
\textbf{Graph-to-3D.} We train the DeepSDF~\cite{park2019deepsdf} version of Graph-to-3D, as this is the only version that can achieve SDF-based generation. Since DeepSDF requires per-category training, for the shape decoder of Graph-to-3D, we train twelve DeepSDF models. We do not train on "floor", as this category includes no meshes and is a virtual node in scene graphs. We first train DeepSDF models following the protocol in the original work \cite{park2019deepsdf} with 1,500 epochs, with the objects that appear in training scenes of SG-FRONT. After training, we optimize the latent code of each training object and store the embeddings. We then train Graph-to-3D with the latent codes following their public training protocol \cite{graph2scene2021}. At the inference time, we use the predicted latent code directly to generate the 3D shapes and, thereby, with the layout, the entire scene. 

\textbf{Graph-to-Box.} It is the layout prediction mode of Graph-to-3D without shape prediction. We remove the shape branch and train it using the same settings as in Graph-to-3D.

\textbf{3D-SLN.} We follow the implementation and training details provided by authors \cite{Luo_2020_CVPR}. We train this baseline for 200 epochs and select the best by the validation accuracy.

\textbf{Progressive.} This is a modified baseline upon Graph-to-Box, specifically adding objects one by one in an autoregressive manner \cite{graph2scene2021}. This can be seen as a method to function in manipulation mode. We train it for 200 epochs and select the best by validation accuracy.

\textbf{Layout+txt2shape.} We utilize the state-of-the-art model SDFusion~\cite{cheng2023sdfusion} as the text-to-shape model and collect all objects in the training scenes for SDFusion to train for 200 epochs using a learning rate of $1e-5$ as proposed in the original paper. Then we train the layout branch solely with the same training settings in our pipeline. Finally, we connect the layout branch with the SDFusion model in series to finish this baseline.

\subsection{Open source artifacts.}
The following open-source artifacts were used in our experiments. We want to thank the respective authors for maintaining an easy-to-use codebase.

\begin{itemize}
    \item \href{https://pytorch3d.org}{PyTorch3D}
    \item \href{https://trimsh.org/trimesh.html}{Trimesh}
    \item \href{http://www.open3d.org}{Open3D}
    \item \href{https://github.com/mseitzer/pytorch-fid}{fid-score}
    \item \href{https://github.com/yccyenchicheng/SDFusion}{SDFusion}
    \item \href{https://github.com/facebookresearch/DeepSDF}{DeepSDF}
    \item \href{https://github.com/he-dhamo/graphto3d}{Graph-to-3D}
    \item \href{https://nv-tlabs.github.io/ATISS/}{ATISS}
\end{itemize}